\pdfoutput=1
\documentclass[11pt]{article}

\usepackage[preprint]{acl}

\usepackage{array}
\usepackage{multirow}
\usepackage{adjustbox}

\usepackage{tabularx} 
\usepackage{geometry} 

\usepackage{caption}      
\usepackage{amssymb}
\usepackage{booktabs} 
\usepackage{times}
\usepackage{latexsym}
\usepackage{amsmath}
\usepackage{natbib}
\usepackage{comment}
\usepackage[T1]{fontenc}
\usepackage[utf8]{inputenc}

\usepackage{microtype}

\usepackage{inconsolata}

\usepackage{graphicx}
\usepackage{kotex}
\title{Investigating Language Preference of Multilingual RAG Systems}

\author{Jeonghyun Park \and Hwanhee Lee\textsuperscript{$\dagger$} \\
    Department of Artificial Intelligence, Chung-Ang University, Seoul, Korea\\
    \texttt{\{tom0365, hwanheelee\}@cau.ac.kr}
}

\begin{document}
\maketitle
\footnotetext{\textsuperscript{$\dagger$}Corresponding author.}

\begin{abstract}
Multilingual Retrieval-Augmented Generation (mRAG) systems enhance language models by integrating external multilingual information to produce context-aware responses. However, mRAG systems struggle with retrieving relevant information due to linguistic variations between queries and documents, generating inconsistent responses when multilingual sources conflict. In this work, we systematically investigate language preferences in both retrieval and generation of mRAG through a series of experiments. Our analysis indicates that retrievers tend to prefer high-resource and query languages, yet this preference does not consistently improve generation performance. Moreover, we observe that generators prefer the query language or Latin scripts, leading to inconsistent outputs. To overcome these issues, we propose Dual Knowledge Multilingual RAG (DKM-RAG), a simple yet effective framework that fuses translated multilingual passages with complementary model knowledge. Empirical results demonstrate that DKM-RAG mitigates language preference in generation and enhances performance across diverse linguistic settings. Code is available at \url{https://github.com/jeonghyunpark2002/LanguagePreference.git}

\end{abstract}

\begin{figure}[t]
    \centering
    \includegraphics[width=1.0\linewidth]{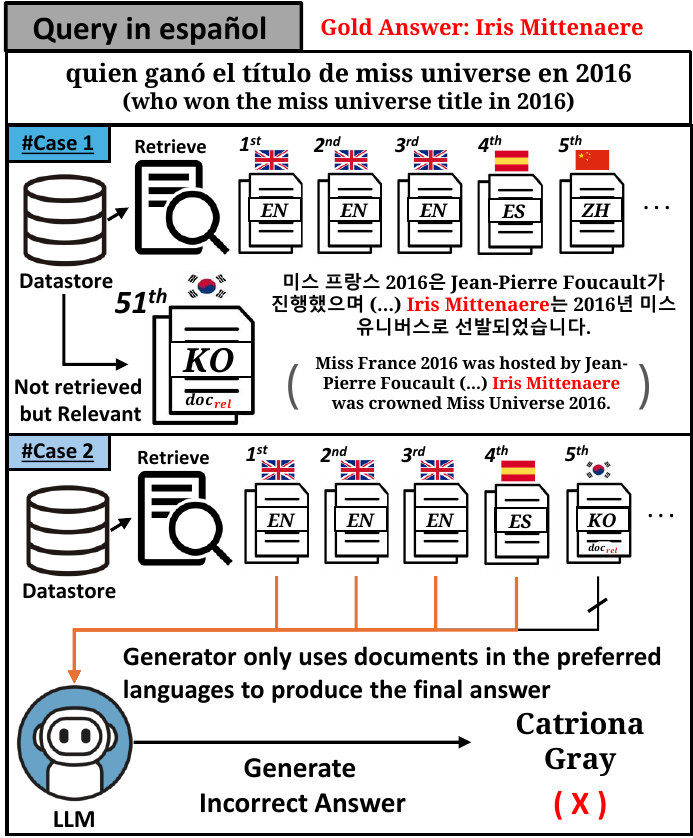}
    \caption{Failure cases of multilingual RAG system showing degraded generation ability because of language preference of retriever and generator in mRAG pipeline. $doc_{rel}$ in the Korean (KO) document represents the relevant document to the given query that can be utilized to generate a final answer.}
    %\vspace{-5mm}
    \label{fig:motivation}
\end{figure}

\section{Introduction}
%mRAG
    
Multilingual Retrieval-Augmented Generation (mRAG) extends traditional Retrieval-Augmented Generation (RAG)~\cite{lewis2020retrieval} by leveraging multilingual external sources to generate accurate, contextually and linguistically aware responses. However, mRAG systems face challenges in retrieving relevant information due to linguistic discrepancies between queries and documents~\cite{wu2024limitscrosslingualdensepassage}. Moreover, conflicts among multilingual sources can lead to inconsistencies in the generated responses~\cite{chataigner2024multilingualhallucinationgapslarge}.

Beyond retrieval challenges and source conflicts, language preference is another critical issue in mRAG systems, often leading to inaccurate outputs. 
As illustrated in \textbf{Case 1} of Figure~\ref{fig:motivation}, the retriever may prioritize particular languages—especially high-resource or query-language documents—at the expense of truly relevant information in low-resource languages. Consequently, the Large Language Model (LLM) either produces an incorrect answer or deems the query unanswerable due to irrelevant content in the documents. Likewise, in \textbf{Case 2}, even if relevant documents are retrieved, the generator might favor passages in the query language or Latin scripts, ignoring essential evidence in lower-resource languages and resulting in inaccurate outputs. These preferences ultimately limit the effectiveness of mRAG, yielding biased rankings, reduced answer quality, and inconsistencies across languages~\cite{sharma2024fauxpolyglotstudyinformation}.

Prior studies~\cite{10.1145/3626772.3657943, 10.1145/3539813.3545131, sharma2024fauxpolyglotstudyinformation} have investigated this issue by introducing language fairness metrics to assess whether documents from different languages are ranked equitably via statistical equivalence testing, by proposing Language-Preference-Based Re-ranking for multilingual information retrieval, and investigating LLM’s linguistic preference in across-language RAG-based information search setting. However, these approaches primarily focus on a limited set of languages and fail to reflect the true ranking dynamics of documents across languages.

In this work, we aim to understand language preference phenomena in mRAG systems comprehensively. We focus on the following three key research questions:

\begin{itemize}
    \item \textbf{RQ1 (\S\ref{sec:retriever}):} \textit{Which languages does the retriever prefer?}
    \item \textbf{RQ2 (\S\ref{sec:generator}):} \textit{Which languages does the generator prefer, and how do these preferences correlate with mRAG performance?}
    \item \textbf{RQ3 (\S\ref{sec:dual}):} \textit{How can we mitigate language preference in mRAG?}
\end{itemize}

To address these questions, we present a comprehensive evaluation of language preferences throughout the entire mRAG pipeline across multiple languages. To systematically investigate the language preference problem of multilingual retrievers, we propose MultiLingualRankShift (MLRS), a novel metric that quantifies language preference at the retriever level by measuring the shift in document rankings when non-query-language passages are translated into the query language. Our extensive experiments with diverse language combinations demonstrate that the retriever strongly prefers documents that are in high-resource languages and also share the same language as the query, confirming the presence of significant preference (\S\ref{sec:retriever}).

At the generator level, we evaluate language preference by generating responses in multiple languages for the same query and the same retrieved document set, measuring their semantic similarity. 
Our results show that the generator favors both query languages and Latin script languages, with a relatively modest preference for query languages. This ultimately results in a decline in answer quality.
Moreover, we uncover a nuanced relationship between language preference and overall mRAG performance. We observe that a strong preference for high-resource languages does not always lead to improved mRAG performance (\S\ref{sec:generator}).
This occurs because the retriever may retrieve high-resource but irrelevant documents so that the generator cannot generate accurate answers from them. Therefore, language preference can degrade performance by overlooking lower-resource but relevant documents, thereby causing inconsistencies.

Finally, we propose Dual Knowledge Multilingual RAG (DKM-RAG), a simple yet effective framework that mitigates the language preference of mRAG. DKM-RAG enhances mRAG by combining externally retrieved, translated passages with internally rewritten passages enriched by the model’s knowledge. Empirical results demonstrate that DKM-RAG significantly reduces language preference issues in the generation process, leading to improved performance across a range of linguistic settings (\S\ref{sec:dual}).

\begin{figure*}[t]
    \centering
    \includegraphics[width=1.0\linewidth]{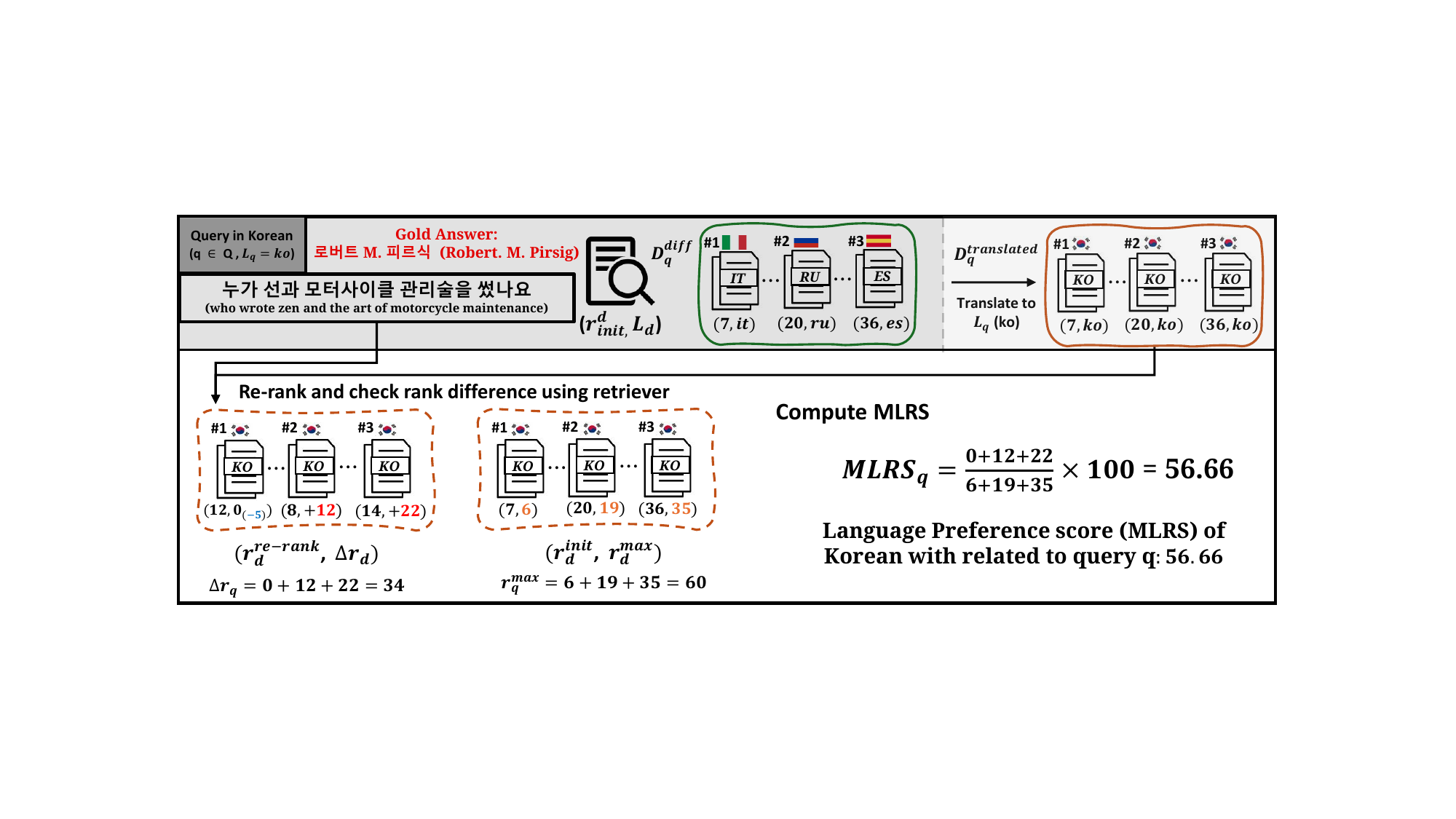}
    \caption{Overall framework of calculating MLRS. For simplicity, we only consider three documents to calculate the MLRS score.}
    %\vspace{-3mm}
    \label{fig:mlr}
\end{figure*}

\section{MultiLingualRankShift}
To evaluate the language preference of a multilingual retriever in the mRAG system, we introduce \textit{MultiLingualRankShift} (MLRS), a novel metric that quantifies how much the ranking of retrieved documents improves when non-query language documents are translated into the query language. As shown in Figure~\ref{fig:mlr}, MLRS is computed in three stages: (i) retrieving documents across multiple languages, (ii) translating documents that are not in the query language into the query language, and (iii) re-ranking the translated documents to measure rank improvements.

\subsection{Stage 1: Initial Document Retrieval}
For each query $q \in Q$ (where $Q$ is the set of all queries), we retrieve a ranked list of documents $D_q$ from multilingual datastores. Each document $d \in D_q$ is assigned an initial rank $r_d^{\text{init}}$ (with 1 being the highest rank). Let $L_q$ denote the language of the query and $L_d$ the language of document $d$. 

\subsection{Stage 2: Translation of Non-Query Language Documents}
To ensure language consistency when assessing ranking improvements, we extract documents whose language differs from that of the query. Formally, we define:
$$
D_q^{\text{diff}} = \{\, (d,\, r_d^{\text{init}}) \mid d \in D_q,\; L_d \neq L_q \,\}.
$$
Each document in $D_q^{\text{diff}}$ is then translated into the query language $L_q$, resulting in the set:
$$
D_q^{\text{Translated}} = \{\, d \mid d \text{ has been translated into } L_q \,\}.
$$

\subsection{Stage 3: Re-Ranking and MLRS Score Computation}
The translated documents in $D_q^{\text{Translated}}$ are re-ranked using retrievers in conjunction with the original query. Let $r_d^{\text{re-rank}}$ denote the new rank of document $d$ after re-ranking. To capture ranking improvements, we compute the rank difference for each document $d$ as:
$$
\Delta r_d = \max\bigl( r_d^{\text{init}} - r_d^{\text{re-rank}},\, 0 \bigr).
$$
A positive value of $\Delta r_d$ indicates that the document has moved up in the ranking. For each query $q$, the total observed improvement is given by:
$$
\Delta r_q = \sum_{d \in D_q^{\text{Translated}}} \Delta r_d.
$$

To normalize this improvement, we first define the maximum possible improvement for each document as:
$$
\Delta r_d^{\max} = r_d^{\text{init}} - 1,
$$
and then compute the total maximum improvement for query $q$:
$$
\Delta r_q^{\max} = \sum_{d \in D_q^{\text{Translated}}} \Delta r_d^{\max}.
$$

The query-specific MLRS score is then calculated as:
$$
\text{MLRS}_q = 
\begin{cases}
\displaystyle \frac{\Delta r_q}{\Delta r_q^{\max}} \times 100, & \text{if } \Delta r_q^{\max} > 0,\\[6pt]
0, & \text{otherwise}.
\end{cases}
$$
Finally, the overall MultiLingualRankShift is obtained by averaging the scores over all queries:
$$
\text{MLRS} = \frac{1}{\lvert Q \rvert} \sum_{q \in Q} \text{MLRS}_q.
$$

%\vspace{-1mm}

\section{General Setup}

\subsection{Dataset}
By following previous study~\cite{chirkova-etal-2024-retrieval}, we use MKQA~\cite{longpre-etal-2021-mkqa} dataset, a multilingual open
domain question answering evaluation set in our experiments. MKQA consists of 10k examples from the Natural Questions (NQ) dataset~\cite{kwiatkowski2019natural}, translated into 25 languages. This dataset is therefore parallel between languages and grounds knowledge primarily in English Wikipedia. In our experiments, we also select a subset of 2.7K samples, overlapping between MKQA and KILT NQ datasets~\footnote{\url{https://huggingface.co/datasets/facebook/kilt_tasks}}, thus recovering relevant documents information from KILT NQ.

\subsection{Models}
\paragraph{Multilingual Retrievers} 
Following previous work~\cite{chirkova-etal-2024-retrieval}, we use a strong and publicly available BGE-m3~\cite{bge-m3} as our multilingual retriever which can encode various languages we consider in our experiments.
Consistent with the retriever, we use BGE-m3~\cite{bge-m3} as a re-ranking encoder for computing MLRS. In addition, we use two other Sentence-BERT series re-ranking encoders~\cite{reimers-2019-sentence-bert}, paraphrase-multilingual-MiniLM-L12-v2 and paraphrase-multilingual-mpnet-base-v2 to generalize the experimental results. We abbreviate them as p-mMiniLM and p-mMpNet for better visibility of the table.

%\vspace{-1mm}
\paragraph{Generators}
 We use a recently released strong multilingual LLM, aya-expanse-8b~\cite{dang2024ayaexpansecombiningresearch} that can deal with various languages well. Also, we use strong LLMs, qwen 2.5-7B Instruct~\cite{qwen2.5} and Phi-4 14B~\cite{abdin2024phi4technicalreport}, Llama-3.1-8B-Instruct~\cite{dubey2024llama} as our generators.
                
\subsection{Other Implementation Details}
\paragraph{Translation model}
We utilize a robust translation model for various languages, NLLB-200-distilled-600M~\cite{costa2022no} in our experiments.

%\vspace{-1mm}
\paragraph{Datastore}
Following previous study~\cite{chirkova-etal-2024-retrieval}, we use Wikipedia as our datastore documents collection.  In most of our experiments, we retrieve from two main Wikipedia sources: the KILT version of English Wikipedia\footnote{\url{https://huggingface.co/datasets/facebook/kilt_wikipedia}} and the Wikipedia edition in the user’s native language\footnote{\url{https://huggingface.co/datasets/wikimedia/wikipedia}}.
For detailed statistics, please refer to Appendix~\ref{appendix:statistics}.

\paragraph{Baseline}
We conduct several experiments to measure the language preference of mRAG based on Bergen~\cite{chirkova-etal-2024-retrieval}. Bergen explores the components and adjustments necessary to develop an effective mRAG pipeline, serving as a robust baseline for future research. 

\begin{table*}[!ht]
\centering
\setlength{\tabcolsep}{3pt}
\renewcommand{\arraystretch}{1.0}
\resizebox{0.95\textwidth}{!}{
\begin{tabular}{cc|>{\columncolor{gray!15}}c|cccccccc}
\toprule
\multicolumn{2}{c|}{} 
& \textbf{\(L_q = L_d\)} 
& \multicolumn{8}{c}{\textbf{\(L_q \neq L_d\)}} \\
\textbf{Query Lang.} & \textbf{Encoder}
& 
& \textbf{en} & \textbf{ko} & \textbf{zh} & \textbf{fr} & \textbf{ja} & \textbf{it} & \textbf{pt} & \textbf{es} \\
\midrule
% =========================
% QL = en
% =========================
\multirow{3}{*}{\textbf{en}}
& \textbf{bge-m3}
  & \textbf{56.03} 
  & -- 
  & 33.02 {\scriptsize(\textcolor{blue}{-23.01})}
  & 33.10 {\scriptsize(\textcolor{blue}{-22.93})}
  & 36.61 {\scriptsize(\textcolor{blue}{-19.42})}
  & 33.36 {\scriptsize(\textcolor{blue}{-22.67})}
  & 35.89 {\scriptsize(\textcolor{blue}{-20.14})}
  & 35.86 {\scriptsize(\textcolor{blue}{-20.17})}
  & \underline{36.62} {\scriptsize(\textcolor{blue}{-19.41})} \\
& \textbf{p-mMiniLM}
  & \textbf{56.85}
  & -- 
  & 34.34 {\scriptsize(\textcolor{blue}{-22.51})}
  & 34.61 {\scriptsize(\textcolor{blue}{-22.24})}
  & \underline{38.17} {\scriptsize(\textcolor{blue}{-18.68})}
  & 34.52 {\scriptsize(\textcolor{blue}{-22.33})}
  & 37.15 {\scriptsize(\textcolor{blue}{-19.70})}
  & 36.73 {\scriptsize(\textcolor{blue}{-20.12})}
  & 37.96 {\scriptsize(\textcolor{blue}{-18.89})} \\
& \textbf{p-mMpNet}
  & \textbf{57.49}
  & -- 
  & 34.45 {\scriptsize(\textcolor{blue}{-23.04})}
  & 34.27 {\scriptsize(\textcolor{blue}{-23.22})}
  & \underline{37.94} {\scriptsize(\textcolor{blue}{-19.55})}
  & 34.67 {\scriptsize(\textcolor{blue}{-22.82})}
  & 37.34 {\scriptsize(\textcolor{blue}{-20.15})}
  & 37.02 {\scriptsize(\textcolor{blue}{-20.47})}
  & 37.90 {\scriptsize(\textcolor{blue}{-19.59})} \\
\midrule
% =========================
% QL = ko
% =========================
\multirow{3}{*}{\textbf{ko}}
& \textbf{bge-m3}
  & \underline{41.15} 
  & \textbf{43.49} {\scriptsize(\textcolor{red}{+2.34})}
  & -- 
  & 34.42 {\scriptsize(\textcolor{blue}{-6.73})}
  & 36.42 {\scriptsize(\textcolor{blue}{-4.73})}
  & 37.18 {\scriptsize(\textcolor{blue}{-3.97})}
  & 35.72 {\scriptsize(\textcolor{blue}{-5.43})}
  & 35.30 {\scriptsize(\textcolor{blue}{-5.85})}
  & 35.93 {\scriptsize(\textcolor{blue}{-5.22})} \\
& \textbf{p-mMiniLM}
  & \underline{42.95}
  & \textbf{44.62} {\scriptsize(\textcolor{red}{+1.67})}
  & -- 
  & 36.04 {\scriptsize(\textcolor{blue}{-6.91})}
  & 37.08 {\scriptsize(\textcolor{blue}{-5.87})}
  & 38.47 {\scriptsize(\textcolor{blue}{-4.48})}
  & 36.07 {\scriptsize(\textcolor{blue}{-6.88})}
  & 36.18 {\scriptsize(\textcolor{blue}{-6.77})}
  & 36.45 {\scriptsize(\textcolor{blue}{-6.50})} \\
& \textbf{p-mMpNet}
  & \underline{42.53}
  & \textbf{44.98} {\scriptsize(\textcolor{red}{+2.45})}
  & -- 
  & 35.85 {\scriptsize(\textcolor{blue}{-6.68})}
  & 37.20 {\scriptsize(\textcolor{blue}{-5.33})}
  & 39.01 {\scriptsize(\textcolor{blue}{-3.52})}
  & 36.21 {\scriptsize(\textcolor{blue}{-6.32})}
  & 35.65 {\scriptsize(\textcolor{blue}{-6.88})}
  & 36.34 {\scriptsize(\textcolor{blue}{-6.19})} \\
\midrule
% =========================
% QL = zh
% =========================
\multirow{3}{*}{\textbf{zh}}
& \textbf{bge-m3}
  & \underline{44.98}
  & \textbf{45.26} {\scriptsize(\textcolor{red}{+0.28})}
  & 34.52 {\scriptsize(\textcolor{blue}{-10.46})}
  & -- 
  & 36.34 {\scriptsize(\textcolor{blue}{-8.64})}
  & 36.05 {\scriptsize(\textcolor{blue}{-8.93})}
  & 35.86 {\scriptsize(\textcolor{blue}{-9.12})}
  & 35.73 {\scriptsize(\textcolor{blue}{-9.25})}
  & 36.45 {\scriptsize(\textcolor{blue}{-8.53})} \\
& \textbf{p-mMiniLM}
  & \textbf{46.18}
  & \underline{45.39} {\scriptsize(\textcolor{blue}{-0.79})}
  & 35.46 {\scriptsize(\textcolor{blue}{-10.72})}
  & -- 
  & 36.98 {\scriptsize(\textcolor{blue}{-9.20})}
  & 36.77 {\scriptsize(\textcolor{blue}{-9.41})}
  & 36.38 {\scriptsize(\textcolor{blue}{-9.80})}
  & 36.05 {\scriptsize(\textcolor{blue}{-10.13})}
  & 36.85 {\scriptsize(\textcolor{blue}{-9.33})} \\
& \textbf{p-mMpNet}
  & \textbf{46.27}
  & \underline{45.41} {\scriptsize(\textcolor{blue}{-0.86})}
  & 35.21 {\scriptsize(\textcolor{blue}{-11.06})}
  & -- 
  & 36.87 {\scriptsize(\textcolor{blue}{-9.40})}
  & 36.71 {\scriptsize(\textcolor{blue}{-9.56})}
  & 36.28 {\scriptsize(\textcolor{blue}{-9.99})}
  & 35.94 {\scriptsize(\textcolor{blue}{-10.33})}
  & 36.78 {\scriptsize(\textcolor{blue}{-9.49})} \\
\midrule
% =========================
% QL = fr
% =========================
\multirow{3}{*}{\textbf{fr}}
& \textbf{bge-m3}
  & \underline{43.18}
  & \textbf{47.23} {\scriptsize(\textcolor{red}{+4.05})}
  & 33.29 {\scriptsize(\textcolor{blue}{-9.89})}
  & 33.58 {\scriptsize(\textcolor{blue}{-9.60})}
  & -- 
  & 34.07 {\scriptsize(\textcolor{blue}{-9.11})}
  & 36.70 {\scriptsize(\textcolor{blue}{-6.48})}
  & 36.30 {\scriptsize(\textcolor{blue}{-6.88})}
  & 37.25 {\scriptsize(\textcolor{blue}{-5.93})} \\
& \textbf{p-mMiniLM}
  & \underline{44.09}
  & \textbf{48.15} {\scriptsize(\textcolor{red}{+4.06})}
  & 34.54 {\scriptsize(\textcolor{blue}{-9.55})}
  & 34.52 {\scriptsize(\textcolor{blue}{-9.57})}
  & -- 
  & 34.83 {\scriptsize(\textcolor{blue}{-9.26})}
  & 37.65 {\scriptsize(\textcolor{blue}{-6.44})}
  & 37.05 {\scriptsize(\textcolor{blue}{-7.04})}
  & 38.03 {\scriptsize(\textcolor{blue}{-6.06})} \\
& \textbf{p-mMpNet}
  & \underline{43.96}
  & \textbf{48.14} {\scriptsize(\textcolor{red}{+4.18})}
  & 34.25 {\scriptsize(\textcolor{blue}{-9.71})}
  & 34.37 {\scriptsize(\textcolor{blue}{-9.59})}
  & -- 
  & 34.61 {\scriptsize(\textcolor{blue}{-9.35})}
  & 37.59 {\scriptsize(\textcolor{blue}{-6.37})}
  & 36.93 {\scriptsize(\textcolor{blue}{-7.03})}
  & 38.01 {\scriptsize(\textcolor{blue}{-5.95})} \\
\midrule
% =========================
% QL = ja
% =========================
\multirow{3}{*}{\textbf{ja}}
& \textbf{bge-m3}
  & \underline{45.03}
  & \textbf{45.18} {\scriptsize(\textcolor{red}{+0.15})}
  & 35.45 {\scriptsize(\textcolor{blue}{-9.58})}
  & 34.86 {\scriptsize(\textcolor{blue}{-10.17})}
  & 36.71 {\scriptsize(\textcolor{blue}{-8.32})}
  & -- 
  & 36.11 {\scriptsize(\textcolor{blue}{-8.92})}
  & 35.88 {\scriptsize(\textcolor{blue}{-9.15})}
  & 36.56 {\scriptsize(\textcolor{blue}{-8.47})} \\
& \textbf{p-mMiniLM}
  & \textbf{45.80}
  & \underline{45.54} {\scriptsize(\textcolor{blue}{-0.26})}
  & 35.90 {\scriptsize(\textcolor{blue}{-9.90})}
  & 35.57 {\scriptsize(\textcolor{blue}{-10.23})}
  & 37.18 {\scriptsize(\textcolor{blue}{-8.62})}
  & -- 
  & 36.53 {\scriptsize(\textcolor{blue}{-9.27})}
  & 36.25 {\scriptsize(\textcolor{blue}{-9.55})}
  & 36.91 {\scriptsize(\textcolor{blue}{-8.89})} \\
& \textbf{p-mMpNet}
  & \textbf{45.67}
  & \underline{45.39} {\scriptsize(\textcolor{blue}{-0.28})}
  & 35.73 {\scriptsize(\textcolor{blue}{-9.94})}
  & 35.30 {\scriptsize(\textcolor{blue}{-10.37})}
  & 36.94 {\scriptsize(\textcolor{blue}{-8.73})}
  & -- 
  & 36.24 {\scriptsize(\textcolor{blue}{-9.43})}
  & 35.98 {\scriptsize(\textcolor{blue}{-9.69})}
  & 36.62 {\scriptsize(\textcolor{blue}{-9.05})} \\
\midrule
% =========================
% QL = it
% =========================
\multirow{3}{*}{\textbf{it}}
& \textbf{bge-m3}
  & \underline{41.06}
  & \textbf{46.63} {\scriptsize(\textcolor{red}{+5.57})}
  & 33.30 {\scriptsize(\textcolor{blue}{-7.76})}
  & 33.47 {\scriptsize(\textcolor{blue}{-7.59})}
  & 37.92 {\scriptsize(\textcolor{blue}{-3.14})}
  & 33.86 {\scriptsize(\textcolor{blue}{-7.20})}
  & -- 
  & 36.44 {\scriptsize(\textcolor{blue}{-4.62})}
  & 37.68 {\scriptsize(\textcolor{blue}{-3.38})} \\
& \textbf{p-mMiniLM}
  & \underline{42.11}
  & \textbf{47.69} {\scriptsize(\textcolor{red}{+5.58})}
  & 34.57 {\scriptsize(\textcolor{blue}{-7.54})}
  & 34.59 {\scriptsize(\textcolor{blue}{-7.52})}
  & 39.07 {\scriptsize(\textcolor{blue}{-3.04})}
  & 34.80 {\scriptsize(\textcolor{blue}{-7.31})}
  & -- 
  & 37.55 {\scriptsize(\textcolor{blue}{-4.56})}
  & 38.83 {\scriptsize(\textcolor{blue}{-3.28})} \\
& \textbf{p-mMpNet}
  & \underline{41.98}
  & \textbf{47.59} {\scriptsize(\textcolor{red}{+5.61})}
  & 34.48 {\scriptsize(\textcolor{blue}{-7.50})}
  & 34.68 {\scriptsize(\textcolor{blue}{-7.30})}
  & 38.94 {\scriptsize(\textcolor{blue}{-3.04})}
  & 34.67 {\scriptsize(\textcolor{blue}{-7.31})}
  & -- 
  & 37.27 {\scriptsize(\textcolor{blue}{-4.71})}
  & 38.67 {\scriptsize(\textcolor{blue}{-3.31})} \\
\midrule
% =========================
% QL = pt
% =========================
\multirow{3}{*}{\textbf{pt}}
& \textbf{bge-m3}
  & \underline{39.19}
  & \textbf{46.64} {\scriptsize(\textcolor{red}{+7.45})}
  & 33.37 {\scriptsize(\textcolor{blue}{-5.82})}
  & 33.46 {\scriptsize(\textcolor{blue}{-5.73})}
  & 37.83 {\scriptsize(\textcolor{blue}{-1.36})}
  & 34.02 {\scriptsize(\textcolor{blue}{-5.17})}
  & 37.13 {\scriptsize(\textcolor{blue}{-2.06})}
  & -- 
  & 38.61 {\scriptsize(\textcolor{blue}{-0.58})} \\
& \textbf{p-mMiniLM}
  & \underline{40.17}
  & \textbf{47.75} {\scriptsize(\textcolor{red}{+7.58})}
  & 34.67 {\scriptsize(\textcolor{blue}{-5.50})}
  & 34.91 {\scriptsize(\textcolor{blue}{-5.26})}
  & 39.02 {\scriptsize(\textcolor{blue}{-1.15})}
  & 35.03 {\scriptsize(\textcolor{blue}{-5.14})}
  & 38.25 {\scriptsize(\textcolor{blue}{-1.92})}
  & -- 
  & 39.68 {\scriptsize(\textcolor{blue}{-0.49})} \\
& \textbf{p-mMpNet}
  & \underline{39.91}
  & \textbf{47.30} {\scriptsize(\textcolor{red}{+7.39})}
  & 34.68 {\scriptsize(\textcolor{blue}{-5.23})}
  & 34.50 {\scriptsize(\textcolor{blue}{-5.41})}
  & 38.70 {\scriptsize(\textcolor{blue}{-1.21})}
  & 34.72 {\scriptsize(\textcolor{blue}{-5.19})}
  & 38.01 {\scriptsize(\textcolor{blue}{-1.90})}
  & -- 
  & 39.35 {\scriptsize(\textcolor{blue}{-0.56})} \\
\midrule
% =========================
% QL = es
% =========================
\multirow{3}{*}{\textbf{es}}
& \textbf{bge-m3}
  & \underline{40.76}
  & \textbf{46.93} {\scriptsize(\textcolor{red}{+6.17})}
  & 33.36 {\scriptsize(\textcolor{blue}{-7.40})}
  & 33.42 {\scriptsize(\textcolor{blue}{-7.34})}
  & 37.73 {\scriptsize(\textcolor{blue}{-3.03})}
  & 33.87 {\scriptsize(\textcolor{blue}{-6.89})}
  & 37.22 {\scriptsize(\textcolor{blue}{-3.54})}
  & 36.88 {\scriptsize(\textcolor{blue}{-3.88})}
  & -- \\
& \textbf{p-mMiniLM}
  & \underline{41.81}
  & \textbf{47.90} {\scriptsize(\textcolor{red}{+6.09})}
  & 34.63 {\scriptsize(\textcolor{blue}{-7.18})}
  & 34.52 {\scriptsize(\textcolor{blue}{-7.29})}
  & 38.86 {\scriptsize(\textcolor{blue}{-2.95})}
  & 34.76 {\scriptsize(\textcolor{blue}{-7.05})}
  & 38.33 {\scriptsize(\textcolor{blue}{-3.48})}
  & 37.84 {\scriptsize(\textcolor{blue}{-3.97})}
  & -- \\
& \textbf{p-mMpNet}
  & \underline{41.33}
  & \textbf{47.34} {\scriptsize(\textcolor{red}{+6.01})}
  & 34.39 {\scriptsize(\textcolor{blue}{-6.94})}
  & 34.19 {\scriptsize(\textcolor{blue}{-7.14})}
  & 38.34 {\scriptsize(\textcolor{blue}{-2.99})}
  & 34.39 {\scriptsize(\textcolor{blue}{-6.94})}
  & 37.73 {\scriptsize(\textcolor{blue}{-3.60})}
  & 37.25 {\scriptsize(\textcolor{blue}{-4.08})}
  & -- \\
\bottomrule
\end{tabular}
}
\begin{comment}
\caption{
Language preference measured by MLR with various re-ranking encoders for various query and document language combinations in a multilingual retriever. The \(L_q=L_d\) column reports the diagonal scores where the query language matches the translated document language, while the remaining columns represent cross-lingual scenarios (i.e., where the query language differs from the document language). Scores in parentheses indicate the difference from the diagonal value (\textcolor{red}{positive} for an improvement, \textcolor{blue}{negative} for a decline). The highest score for each row is highlighted in bold, and the second highest is underlined.
}
\end{comment}
\caption{Language preference measured by MLRS with different re-ranking encoders for various query–document language pairs. The \(L_q=L_d\) column shows scores for matching query and document languages, while the remaining columns represent cross-lingual scenarios. Parentheses indicate the change from the \(L_q=L_d\) column (\textcolor{red}{positive} for improvement, \textcolor{blue}{negative} for decline). The highest score per row is in bold, and the second highest is underlined.}
%\vspace{-2mm}
\label{tab:subset_mlr}
\end{table*}

\section{Language Preference of Retrievers}
\label{sec:retriever}

In this section, we examine two factors that may affect the retriever's language preference: (i) the relationship between the query language ($L_q$) and the document language ($L_d$), and (ii) the resource availability of the languages involved.

%%%%%%%%%%%%%%%%%%%%%%%%%%%%%%%%%%%%%%%%%%%%%%%%%%%%%%%%%%%%%%%%%%%%%
\subsection{Effect of the $L_q$ and $L_d$ Relationship}
%%%%%%%%%%%%%%%%%%%%%%%%%%%%%%%%%%%%%%%%%%%%%%%%%%%%%%%%%%%%%%%%%%%%%
\subsubsection{Experimental Setup}
We evaluate eight language pairs under two scenarios: (1) \textbf{monolingual settings} where $L_q = L_d$, and (2) \textbf{cross-lingual settings} where $L_q \neq L_d$. In each case, our primary metric is \textit{MLRS} (MultiLingualRankShift), computed using three re-ranker encoders (bge-m3, p-mMiniLM, and p-mMpNet). For example, if the query is in English (en) but the target translation is in Korean (ko), we translate all non-English passages into Korean and then measure the rank changes with MLRS.

\subsubsection{Results for Monolingual Settings ($L_q = L_d$)}
\paragraph{Strong Preference When the Query and Document Languages Match.}  
As shown in the leftmost column of Table~\ref{tab:subset_mlr}, when the query and document languages are identical, the retriever shows a high preference. This is expected, as direct linguistic alignment avoids the complexities of cross-lingual mapping and translation, thereby yielding stronger preference.

\subsubsection{Results for Cross-Lingual Settings ($L_q \neq L_d$)}
\paragraph{Lower Overall MLRS in Cross-Lingual Matching.}  
When $L_q \neq L_d$, the retriever performs cross-lingual matching, which typically results in lower MLRS values than in monolingual cases. As indicated by the right-hand columns in Table~\ref{tab:subset_mlr} (highlighted in blue), cross-lingual setups are generally less preferred than their monolingual counterparts—except in cases involving English.

%\vspace{-2mm}
\paragraph{English as a Dominant Target Language.}  
We observe that when the translated document language $L_d$ is English, the retriever exhibits nearly the highest language preference as stated in the English column (en) in Table~\ref{tab:subset_mlr}. In fact, English often outperforms even monolingual configurations, likely due to the abundance of English data in pre-training, which biases the model towards stronger English representations.

%\vspace{-2mm}
\paragraph{Influence of Language Family Similarity.}  
Language family and geographic proximity also play a role. For example, Romance languages (fr, it, pt, es) share extensive lexical and structural similarities, which help maintain a relatively high cross-lingual preference and narrow the performance gap with monolingual setups, as illustrated by the joint $L_q$ and $L_d$ pairs in Table~\ref{tab:subset_mlr}. Similarly, East Asian languages (ko, ja, zh) tend to show moderate declines in cross-lingual scenarios compared to the monolingual baseline, although they still lag behind the highest scores.

%%%%%%%%%%%%%%%%%%%%%%%%%%%%%%%%%%%%%%%%%%%%%%%%%%%%%%%%%%%%%%%%%%%%%
\subsection{Impact of Language Resource Availability}
%%%%%%%%%%%%%%%%%%%%%%%%%%%%%%%%%%%%%%%%%%%%%%%%%%%%%%%%%%%%%%%%%%%%%
\subsubsection{Experimental Setup}
We also investigate whether the volume of available language resources affects MLRS. We categorize languages into three groups based on their distribution in the pre-training corpus of recent LLMs: high-resource (e.g., English), mid-resource (e.g., Spanish), and low-resource (e.g., Korean). We use the same query set across all setups while systematically varying $L_q$ and $L_d$.

%%%%%%%%%%%%%%%%%%%%%%%%%%%%%%%%%%%%%%%%%%%%%%%%%%%%%%%%%%%%%%%%%%%%%
\begin{figure*}[ht]
  \centering
  \includegraphics[width=0.3\textwidth]{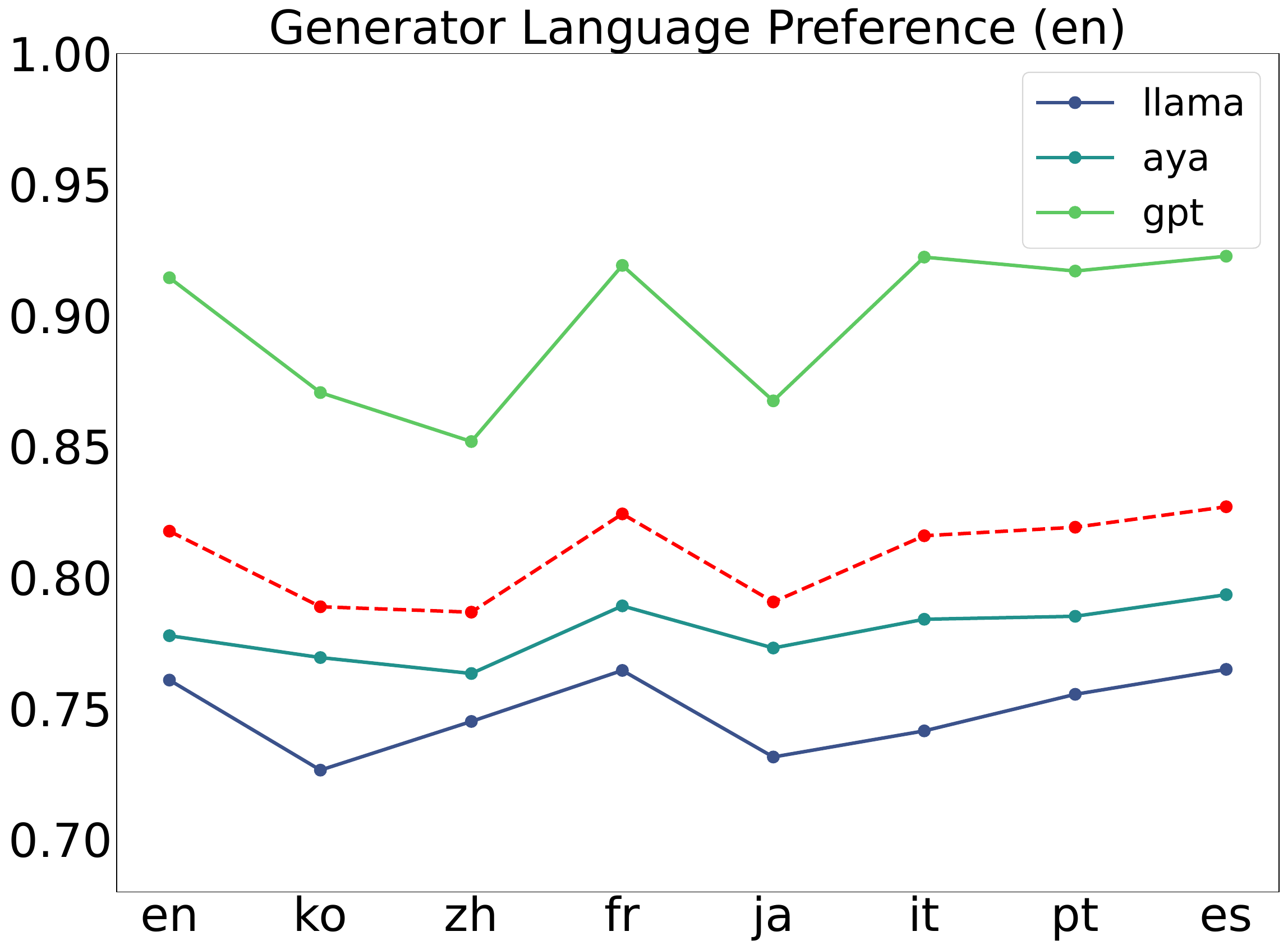}\hfill
  \includegraphics[width=0.3\textwidth]{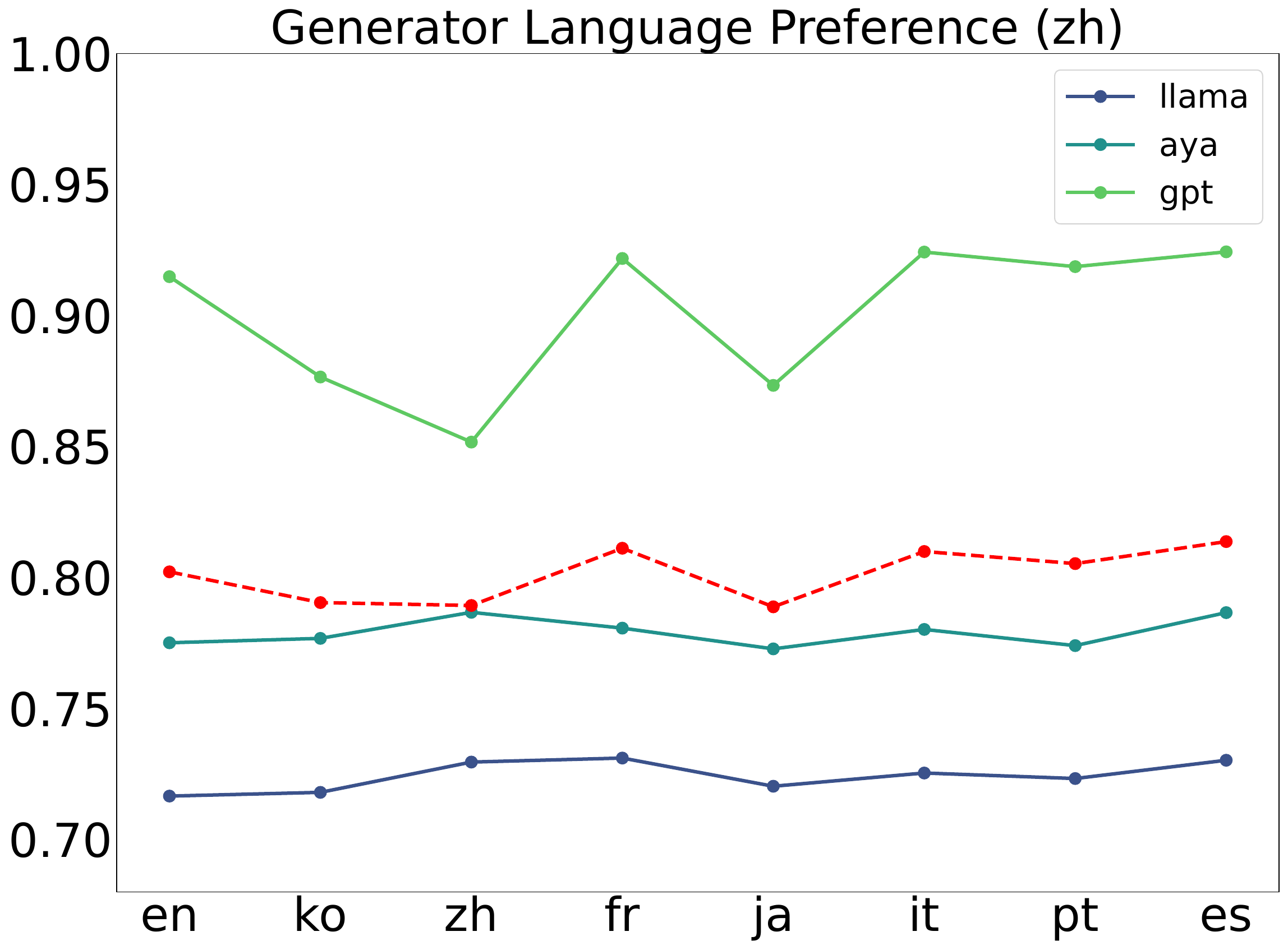}\hfill
  \includegraphics[width=0.3\textwidth]{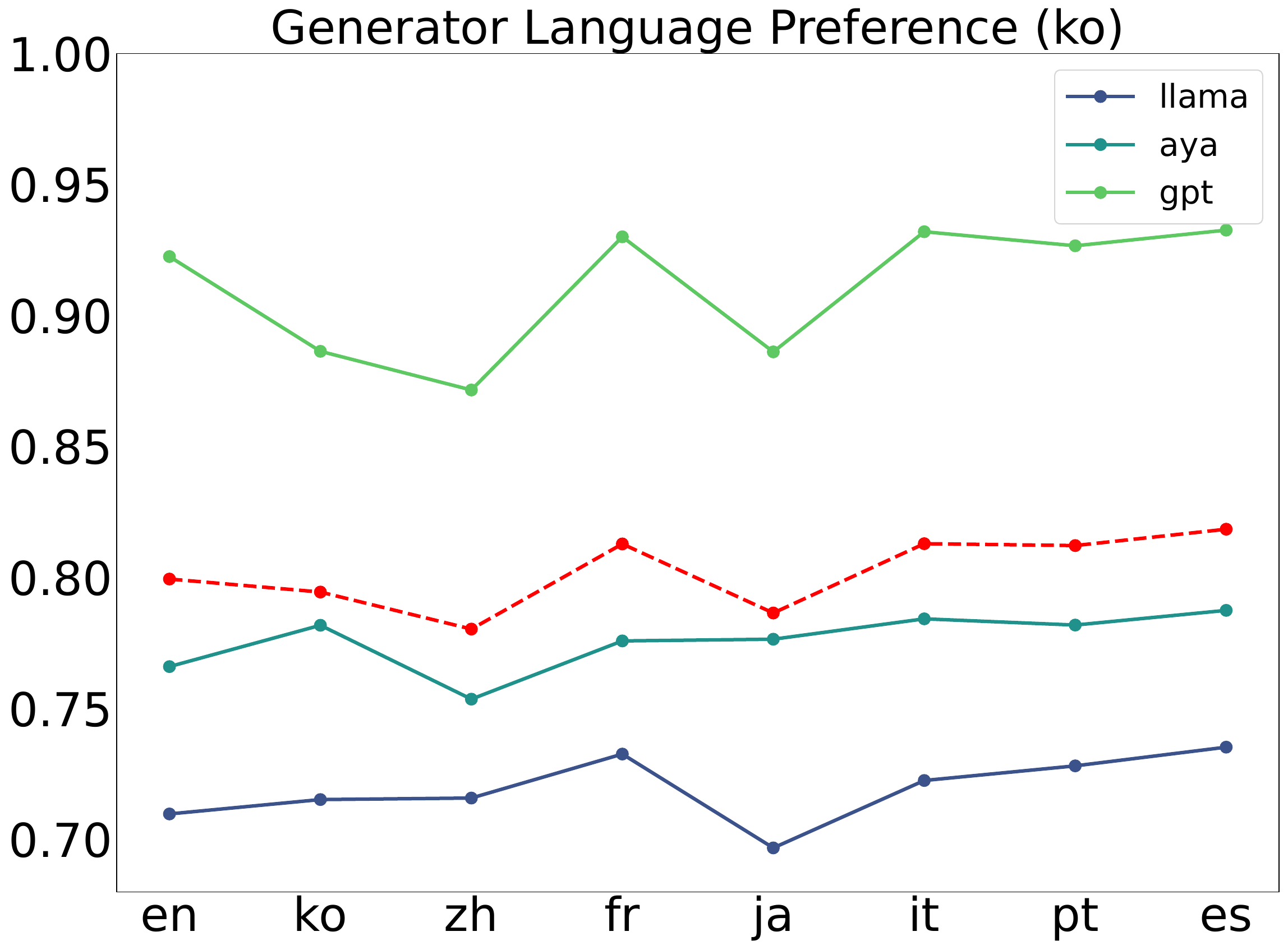}
  \caption{Language Preference of the Generators. In each figure, "aya" represents aya-expanse-8B, "llama" represents Llama-3.1-8B-Instruct, and "gpt" represents gpt-4o-mini. The red dotted line indicates the average generator preference.}
  \label{fig:multi_lang}
  %\vspace{-2mm}
\end{figure*}
%%%%%%%%%%%%%%%%%%%%%%%%%%%%%%%%%%%%%%%%%%%%%%%%%%%%%%%%%%%%%%%%%%%%%

\subsubsection{Results}
\paragraph{Limited Impact of Query Language Resources.}  
The resource level of the query language ($L_q$) has a limited effect on cross-lingual preference. As shown in Table~\ref{tab:subset_mlr}, when $L_q$ is high-resource (e.g., English), strong preference is observed only if $L_d$ also matches a high-resource language. Otherwise, the MLRS scores remain similar regardless of whether $L_q$ is high-, mid-, or low-resource.

\paragraph{Document Language Resources Are More Influential.}  
In contrast, the language resource level of the document language ($L_d$) has a pronounced impact on MLRS. As shown in Table~\ref{tab:subset_mlr}, documents from high-resource languages consistently achieve the highest preference scores, followed by mid-resource and then low-resource languages. This trend (High > Mid > Low) suggests that extensive pre-training on high-resource languages enables stronger alignment, yielding higher MLRS even across diverse query languages. Conversely, low-resource datastores typically produce lower MLRS scores unless the query language also corresponds to that low-resource setting.

Overall, our results indicate that the resource availability of $L_d$ critically influences the language preference of the retriever. These findings lay the groundwork for further investigation into the language dynamics within mRAG systems.

\subsection{Impact of Translation Quality}

To investigate the effect of translation quality on language preference as measured by the MLRS metric, we conduct a small-scale experiment by randomly sampling 100 queries out of the full set of 2,827. We translate these queries using GPT-4o-mini, which demonstrated highly robust translation quality. We evaluate five languages (English, Korean, Spanish, Chinese, and French) and re-ranked the top-10 retrieved documents using the BGE-M3 encoder, and compute rank shifts within the top-10 to derive MLRS for monolingual ($L_q = L_d$) and cross-lingual ($L_q \neq L_d$) settings. The results, shown in Table~\ref{tab:mlrs_translation}, are consistent with prior MLRS findings and further reveal that GPT-based translation amplifies the distinctness of language preference trends, as indicated by larger inter-language preference gaps.

\begin{table}[!ht]
\centering
\setlength{\tabcolsep}{3pt}
\renewcommand{\arraystretch}{1.0}
\resizebox{\columnwidth}{!}{
  \begin{tabular}{c|>{\columncolor{gray!15}}c|ccccc}
    \toprule
    \multicolumn{1}{c|}{} 
      & \multicolumn{1}{>{\columncolor{gray!15}}c|}{\textbf{\(L_q = L_d\)}} 
      & \multicolumn{5}{c}{\textbf{\(L_q \neq L_d\)}} \\
    \textbf{Query Lang.} 
      &  
      & \textbf{en} & \textbf{ko} & \textbf{zh} & \textbf{fr} & \textbf{es} \\
    \midrule
    \textbf{en} 
      & \textbf{58.29} 
      & -- 
      & 34.98 {\scriptsize(\textcolor{blue}{-23.31})} 
      & 36.10 {\scriptsize(\textcolor{blue}{-22.19})} 
      & 38.49 {\scriptsize(\textcolor{blue}{-19.80})} 
      & \underline{40.07} {\scriptsize(\textcolor{blue}{-18.22})} \\
    \textbf{ko} 
      & \textbf{47.15} 
      & \underline{42.59} {\scriptsize(\textcolor{blue}{-4.56})} 
      & -- 
      & 37.50 {\scriptsize(\textcolor{blue}{-9.65})} 
      & 39.31 {\scriptsize(\textcolor{blue}{-7.84})} 
      & 38.74 {\scriptsize(\textcolor{blue}{-8.41})} \\
    \textbf{zh} 
      & \textbf{47.42} 
      & \underline{46.66} {\scriptsize(\textcolor{blue}{-0.76})} 
      & 36.89 {\scriptsize(\textcolor{blue}{-10.53})} 
      & -- 
      & 38.24 {\scriptsize(\textcolor{blue}{-9.18})} 
      & 38.10 {\scriptsize(\textcolor{blue}{-9.32})} \\
    \textbf{fr} 
      & \underline{47.35}
      & \textbf{50.76} {\scriptsize(\textcolor{red}{+3.41})} 
      & 34.90 {\scriptsize(\textcolor{blue}{-12.45})} 
      & 36.10 {\scriptsize(\textcolor{blue}{-11.25})} 
      & -- 
      & 38.90 {\scriptsize(\textcolor{blue}{-8.45})} \\
    \textbf{es} 
      & \underline{45.98} 
      & \textbf{50.73} {\scriptsize(\textcolor{red}{+4.75})} 
      & 35.06 {\scriptsize(\textcolor{blue}{-10.92})} 
      & 36.18 {\scriptsize(\textcolor{blue}{-9.80})} 
      & 40.52 {\scriptsize(\textcolor{blue}{-5.46})} 
      & -- \\
    \bottomrule
  \end{tabular}
}
\caption{MLRS scores for monolingual (\(L_q=L_d\)) and cross‐lingual (\(L_q\neq L_d\)) settings on 100 randomly sampled queries.}
\label{tab:mlrs_translation}
\end{table}

\section{Language Preference of Generators}

\label{sec:generator}
In this section, we explore LLM generators' language preferences in mRAG and their impact on overall performance.

%%%%%%%%%%%%%%%%%%%%%%%%%%%%%%%%%%%%%%%%%%%%%%%%%%%%%%%%%%%%%%%%%%%%%
\subsection{Do LLMs Prefer Certain Languages for Contextual Knowledge?}
%%%%%%%%%%%%%%%%%%%%%%%%%%%%%%%%%%%%%%%%%%%%%%%%%%%%%%%%%%%%%%%%%%%%%
\subsubsection{Experimental Setup}
To assess the generator’s language preference, we measure multilingual answer consistency across eight languages: English (en), Korean (ko), Chinese (zh), French (fr), Japanese (ja), Italian (it), Portuguese (pt), and Spanish (es). As shown in Table~\ref{tab:case_generation}, for a given query the generator produces responses in each language using the same set of retrieved documents from the multilingual datastore. We then compute the embedding similarity between each pair of generated answers, resulting in an 8×8 similarity matrix. We define the preference for a specific language as the average similarity score of the responses in that language.

We use LaBSE for measuring multilingual semantic similarity \cite{feng2022language}. And we use aya-expanse-8B, Llama-3.1-8B-Instruct, and GPT-4o-mini as our generators. We use language-specific prompts that incorporate the retrieved passages to induce responses in the target language, enabling us to capture the generator's inherent language preference.

\subsubsection{Results}
\paragraph{Strong Preference for Latin Script Languages.}
Figure~\ref{fig:multi_lang} indicates that the generator produces more consistent responses in languages that use Latin scripts (e.g., en, fr, it, pt, es) compared to non-Latin languages (e.g., ko, zh, ja). This suggests that the model benefits from structural advantages in token alignment when processing Latin-based languages. 

\paragraph{Modest Preference for the Query Languages.}
In addition, the generator shows a slight increase in consistency when the output language matches a query language. For instance, Korean (ko) queries yield somewhat more consistent responses when the generator replies in Korean rather than when the query is in English. However, this improvement is marginal, suggesting that the overall preference toward Latin scripts remains influential. Despite the modest gain, the model still demonstrates some capacity to handle multilingual queries effectively, indicating that matching the query language can provide a small but measurable benefit in non-Latin contexts.

%%%%%%%%%%%%%%%%%%%%%%%%%%%%%%%%%%%%%%%%%%%%%%%%%%%%%%%%%%%%%%%%%%%%%
\subsection{Correlation between Language Preference and mRAG Performance}
%%%%%%%%%%%%%%%%%%%%%%%%%%%%%%%%%%%%%%%%%%%%%%%%%%%%%%%%%%%%%%%%%%%%%
\subsubsection{Experimental Setup}
To examine how language preference impacts overall mRAG performance, we isolate language effects by providing generators with retrieved passages unified in a single target language—selected from the eight candidates (en, ko, zh, fr, ja, it, pt, es). We retrieve data from multilingual sources, enabling a direct comparison between language preference (measured by MLRS, as shown in Table~\ref{tab:subset_mlr}) and performance across three query languages (en, ko, zh).

We evaluate four generators (aya-expanse-8B, Phi-4, Qwen2.5-7B-Instruct, and Llama3.1-8B-Instruct) using character 3-gram recall \cite{chirkova-etal-2024-retrieval} under three configurations. In the first configuration, passages are retrieved from multilingual resources, denoted as \textit{all}. In the second, all retrieved passages are unified into a single target language (single-language document set). Finally, in the third configuration, we employ our proposed DKM-RAG framework (detailed in Section~\ref{sec:dual}) to mitigate language preference. We also compute the average MLRS score (across different query languages) for each language to indicate its overall preference.

\begin{table*}[t]
\centering
\footnotesize	
\renewcommand{\arraystretch}{1.1}
\setlength{\tabcolsep}{5pt}
\begin{tabular}{lcccccccccc}
\toprule
 & \textbf{all} & \textbf{en} & \textbf{zh} & \textbf{ko} & \textbf{fr} & \textbf{ja} & \textbf{it} & \textbf{pt} & \textbf{es} & \textbf{DKM-RAG} \\
\midrule
\multicolumn{11}{l}{\textbf{$L_q =$ en}} \\
\midrule
\textbf{aya-expanse-8b}       & \underline{80.09} & \cellcolor{yellow}79.34 & 63.08 & 64.46 & 76.13 & 61.20 & 75.47 & 75.65 & 76.32 & \textbf{82.60} \\
\textbf{Phi-4}                & \underline{79.69} & \cellcolor{yellow}78.89 & 63.06 & 52.30 & 74.43 & 48.86 & 74.02 & 74.39 & 75.32 & \textbf{82.59} \\
\textbf{Qwen2.5-7B-Instruct}  & \underline{80.15} & \cellcolor{yellow}79.11 & 50.31 & 64.90 & 76.28 & 62.62 & 75.47 & 75.97 & 76.54 & \textbf{82.60} \\
\textbf{Llama3.1-8B-Instruct} & \underline{80.25} & \cellcolor{yellow}79.28 & 61.99 & 65.81 & 76.40 & 62.58 & 75.89 & 76.09 & 76.47 & \textbf{82.57} \\
\midrule
\multicolumn{11}{l}{\textbf{$L_q =$ zh}} \\
\midrule
\textbf{aya-expanse-8b}       & 32.55 & 25.62 & \cellcolor{yellow}\underline{38.31} & 26.64 & 24.00 & 25.27 & 23.63 & 23.63 & 23.79 & \textbf{44.57} \\
\textbf{Phi-4}                & 16.75 & 17.57 & \cellcolor{yellow}\underline{36.76} & 17.50 & 18.15 & 17.56 & 18.19 & 17.89 & 18.44 & \textbf{44.56} \\
\textbf{Qwen2.5-7B-Instruct}  & 34.28 & 27.33 & \cellcolor{yellow}\underline{38.31} & 27.91 & 25.15 & 27.78 & 25.90 & 25.37 & 25.30 & \textbf{44.70} \\
\textbf{Llama3.1-8B-Instruct} & 28.50 & 24.36 & \cellcolor{yellow}\underline{38.48} & 23.84 & 22.48 & 23.78 & 23.18 & 23.32 & 23.02 & \textbf{44.51} \\
\midrule
\multicolumn{11}{l}{\textbf{$L_q =$ ko}} \\
\midrule
\textbf{aya-expanse-8b}       & 40.60 & 38.08 & 26.01 & \cellcolor{yellow}\underline{49.66} & 25.37 & 26.82 & 24.98 & 25.26 & 25.51 & \textbf{55.01} \\
\textbf{Phi-4}                & 26.80 & 20.24 & 17.54 & \cellcolor{yellow}\underline{49.25} & 19.03 & 17.91 & 18.93 & 19.19 & 19.19 & \textbf{54.82} \\
\textbf{Qwen2.5-7B-Instruct}  & 36.50 & 22.87 & 20.08 & \cellcolor{yellow}\underline{49.44} & 21.79 & 20.94 & 21.65 & 21.44 & 21.52 & \textbf{54.85} \\
\textbf{Llama3.1-8B-Instruct} & 37.18 & 26.48 & 22.88 & \cellcolor{yellow}\underline{49.87} & 24.46 & 24.86 & 25.23 & 24.87 & 25.22 & \textbf{54.99} \\
\midrule
\textbf{MLRS (Preference)}     & - & \textbf{47.70} & 35.90 & 35.47 & 37.94 & 37.59 & 37.66 & 37.15 & \underline{37.97} & - \\
\bottomrule
\end{tabular}
\caption{Performance comparison between DKM-RAG and single/all language retrieval settings, showing character 3-gram recall scores for three query languages ($L_q \in \{\text{en}, \text{ko}, \text{zh}\}$) and eight passage languages. The bottom row shows average preference (MLRS) scores. We highlight the cells corresponding to matching query and passage languages with a yellow background. The highest score per row is in bold, and the second highest is underlined.}
%\vspace{-5mm}
\label{tab:resource_comparison_updated}
\end{table*}

\subsubsection{Results and Analysis}
\paragraph{Strong Correlation for English Queries.}
As stated in Table~\ref{tab:resource_comparison_updated}, for queries in English (\( L_q = \text{en} \)), RAG performance shows a strong correlation with language preference. English achieves the best results—likely due to its high-resource availability and the model’s familiarity with it. In this setting, the \textit{all} strategy is particularly effective, as it leverages cross-lingual knowledge fusion. We observe an exception for Japanese (\( \text{ja} \)), where performance is lower despite moderate preference, possibly due to challenges with non-Latin scripts and complex morphology.

\paragraph{Weaker Correlation for Non-English Queries.}
When \( L_q \neq \text{en} \), the relationship between language preference and performance becomes less pronounced. Although the generator generally prefers English passages overall, it achieves optimal performance when it receives retrieved passages that directly match the query language. In these cases, translating all passages into English does not enhance performance; instead, maintaining language consistency between the query and passages yields better results. This finding underscores the importance of linguistic compatibility in mRAG systems.

\paragraph{Optimal mRAG Strategy.}
Based on our experiments, different strategies depending on the query language prove more effective. As stated in Table~\ref{tab:resource_comparison_updated}, for English queries, employing the \textit{all} strategy capitalizes on the high cross-lingual preference for English. In contrast, for non-English queries, translating retrieved passages into the query language \(L_q\) bridges the comprehension gap and ensures better alignment between query intent, passage semantics, and output language. This targeted approach ultimately leads to improved RAG performance by accommodating the specific language dynamics of the generator.

\begin{figure}[h]
    \centering
    \includegraphics[width=1.0\linewidth]{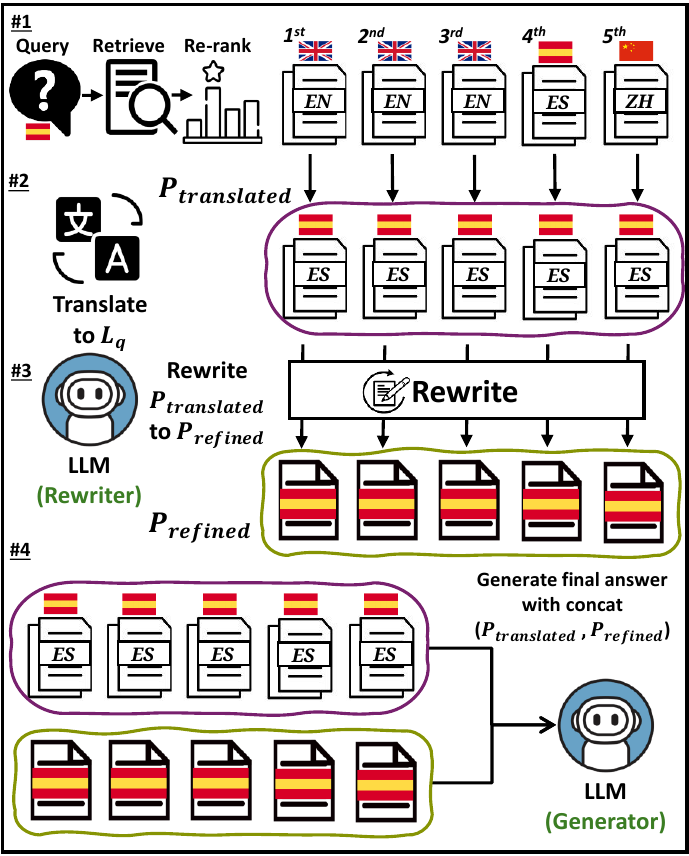}
    \caption{Overall flow of proposed DKM-RAG.}
    %\vspace{-5mm}
    \label{fig:dkm_rag}
\end{figure}

\section{Dual Knowledge Multilingual RAG}
\label{sec:dual}
%\vspace{-1mm}
Translating retrieved documents into the query language benefits mRAG, but it may also reflect retrieval outputs from high-resource languages including irrelevant content. Therefore, leveraging the LLM’s internal knowledge can help filter inaccuracies and enrich the retrieved information with more reliable content. So we rewrite translated passages to refine the relevancy of documents by leveraging LLM's internal information.

Based on this insight, we propose Dual Knowledge Multilingual RAG (DKM-RAG), a framework that leverages both external translated passages and internal knowledge as shown in Figure~\ref{fig:dkm_rag}. First \textbf{(\#1)}, we retrieve documents for a given query from the \textit{all} strategy and re-rank them. Next \textbf{(\#2)}, we obtain external translated passages, \(P_{\text{translated}}\) by translating into the query language. And \textbf{(\#3)}, the rewriter LLM refines each translated passage in the context of the given query to produce refined passages, \(P_{\text{refined}}\). This refining process utilizes a prompt to guide the model in integrating its internal knowledge, removing redundancy, and highlighting relevant information in a coherent and consistent style. For detailed prompts, please refer to Appendix~\ref{appendix:prompts}. Finally \textbf{(\#4)}, We concatenate the two sets to form the final passage set as input to the generator LLM, ensuring that responses are both contextually enriched and linguistically aligned with the query.

\paragraph{Results.} As shown in Table~\ref{tab:resource_comparison_updated}, DKM-RAG outperforms other document-based generator settings. For non-English queries (\(L_{q} \neq \text{en}\)), it leverages translated passages and enriched content to handle linguistic diversity. Even for English queries (\(L_{q} = \text{en}\)), 
it surpasses the \textit{all} baseline, highlighting the importance of integrating translated and refined knowledge.

\paragraph{Ablation Study.}

To prove the effectiveness of concatenating translated passages and refined passages in the DKM-RAG framework, we provide an ablation study of each component in DKM-RAG. As stated in Table~\ref{tab:ablation_dkm}, removing any component from DKM-RAG decreases performance, highlighting that every part is crucial to its effectiveness.

\section{Related Works}

\subsection{Multilingual RAG}
Researchers explore challenges in mRAG, such as the problem of cross-lingual dense passage retrieval for low-resource languages~\cite{wu2024limitscrosslingualdensepassage}, and propose various techniques to address key challenges in mRAG, such as enhancing the performance of language models in low-resource languages~\cite{deshpande2024chain}, resolving low-resource scenarios~\cite{zhang-etal-2024-enhancing-multilingual}, and adapting language models for multilingual reasoning tasks~\cite{yoon-etal-2024-langbridge}. Benchmarks like \textsc{MMTEB}~\cite{enevoldsen2025mmteb} enable systematic evaluation of multilingual retrieval.

Earlier mRAG systems frequently focus on high-resource languages (e.g., English), but a growing body of research aims to make advanced Natural Language Processing (NLP) technology accessible across a wide spectrum of linguistic contexts. Proposed solutions include code-mixed prompts for in-context learning~\cite{shankar-etal-2024-context} and self-distillation from resource-rich to low-resource languages~\cite{zhang-etal-2024-enhancing-multilingual}.

\subsection{Language Preference in mRAG}
Despite significant progress, language preference—a systematic tendency to favor certain languages—remains a critical issue in mRAG systems. This preference arises from imbalances in training data, tokenization mismatches, script differences, and uneven resource availability~\cite{sharma2024fauxpolyglotstudyinformation, wu2024languagesequalinsightsmultilingual}. Studies show that high-resource languages (e.g., English) often overshadow relevant content in lower-resource languages during retrieval~\cite{yang-etal-2024-language-bias, chirkova-etal-2024-retrieval}, leading to suboptimal evidence retrieval~\cite{10.1145/3626772.3657943} and causing inconsistencies or hallucinations in outputs~\cite{chataigner2024multilingualhallucinationgapslarge}. These disparities also raise broader fairness concerns in multilingual NLP, as pre-trained models exhibit group fairness issues across languages~\cite{cabello-piqueras-sogaard-2022-pretrained, ramesh-etal-2023-fairness}.

Researchers propose several methods to counteract language preferences, including language-preference-based re-ranking~\cite{10.1145/3539813.3545131}, evaluate knowledge consistency across languages~\cite{qi-etal-2023-cross}, and specialized datasets designed to detect such imbalances~\cite{li-etal-2024-bordirlines}. However, these approaches often focus on a single mRAG stage or overlook the actual ranking of retrieved documents~\cite{sharma2024fauxpolyglotstudyinformation, 10.1145/3626772.3657943}. We introduce a metric that quantifies language preference in retrieval via ranking differences and propose a simple framework to mitigate these preferences across the entire mRAG pipeline.

\begin{table}[t]
\centering
\renewcommand{\arraystretch}{1.1}
\setlength{\tabcolsep}{6pt} % 열 간격을 약간 줄임
\resizebox{\linewidth}{!}{%
  \begin{tabular}{lccc}
    \toprule
     & \textbf{DKM-RAG} & \textbf{w/o $P_{\text{refined}}$} & \textbf{w/o $P_{\text{translated}}$} \\
    \midrule
    \multicolumn{4}{l}{\textbf{$L_q =$ en}} \\
    \midrule
    \textbf{aya-expanse-8b}       & 82.60 & 79.34 & 81.10 \\
    \textbf{Phi-4}                & 82.59 & 78.89 & 81.08 \\
    \textbf{Qwen2.5-7B-Instruct}  & 82.60 & 79.11 & 81.06 \\
    \textbf{Llama3.1-8B-Instruct} & 82.57 & 79.28 & 81.19 \\
    \midrule
    \multicolumn{4}{l}{\textbf{$L_q =$ zh}} \\
    \midrule
    \textbf{aya-expanse-8b}       & 44.57 & 38.31 & 39.44 \\
    \textbf{Phi-4}                & 44.56 & 36.76 & 38.95 \\
    \textbf{Qwen2.5-7B-Instruct}  & 44.70 & 38.31 & 39.78 \\
    \textbf{Llama3.1-8B-Instruct} & 44.51 & 38.48 & 39.35 \\
    \midrule
    \multicolumn{4}{l}{\textbf{$L_q =$ ko}} \\
    \midrule
    \textbf{aya-expanse-8b}       & 55.01 & 49.66 & 46.15 \\
    \textbf{Phi-4}                & 54.82 & 49.25 & 45.24 \\
    \textbf{Qwen2.5-7B-Instruct}  & 54.85 & 49.44 & 45.32 \\
    \textbf{Llama3.1-8B-Instruct} & 54.99 & 49.87 & 45.55 \\
    \bottomrule
  \end{tabular}%
}
\caption{Ablation study on DKM-RAG. ``DKM-RAG'' denotes the DKM-RAG setting (i.e., the DKM-RAG column in Table~\ref{tab:resource_comparison_updated}), ``w/o $P_{\text{refined}}$'' indicates the performance corresponding to the highlighted cells, and ``w/o $P_{\text{translated}}$'' represents the results using only refined passages.}
\label{tab:ablation_dkm}
\end{table}

\section{Conclusion}
\vspace{-1mm}
In this work, we investigate language preferences in mRAG systems. We propose a metric that measures the language preference of retrievers by checking the rank difference between the translated passage and the original one. Our experiments reveal that retrievers prefer high-resource and query language but do not always yield better generation performance. We also find that generators often favor the query language or Latin scripts, resulting in inconsistent outputs. To address this, we propose DKM-RAG which integrates translated passages with internal knowledge. Empirical results show that DKM-RAG consistently enhances mRAG performance across diverse languages.

\section*{Limitations}
Our approach involves translating documents to measure rank shifts and unify linguistic representations. This process relies heavily on the quality of the translation model employed. Errors or inaccuracies in translation can distort the original meaning of passages and potentially introduce noise into both the retrieval and generation stages. 

MLRS entails translation and re-ranking steps. While this approach offers a principled way to quantify language preference, it also adds latency and computational cost, especially when dealing with large-scale multilingual corpora or real-time systems.

DKM-RAG framework which combines external translated passages and parametric (internal) knowledge, improves performance yet remains relatively straightforward. Future work could explore more sophisticated techniques for merging external and internal knowledge (e.g., trainable fusion mechanisms, dynamic weighting) to further reduce preferences and enhance overall system capabilities.

Lastly, our experiments focus on Wikipedia-based datasets in a specific set of languages, which may not generalize to all linguistic varieties or specialized domains. Future research should examine broader contexts, including low-resource languages not present in widely available corpora or domain-specific retrieval settings, to fully assess how language preferences manifest across diverse real-world scenarios.

\section*{Ethics Statement}
We conduct our experiments using publicly available, multilingual dataset and models that follow recognized research and data-sharing guidelines. These resources are widely utilized in the academic community and are distributed with the intent to minimize harmful biases, inappropriate content, or stereotypes. However, they may still not fully represent the diversity of all languages and cultural contexts. We adhere strictly to the usage protocols and license agreements set forth by the original providers, who have taken steps to ensure compliance with established ethical standards.

\section*{Acknowledgments}
We would like to thank Byeongjeong Kim for his comments and feedback about our figures. We also thank Gyutae Park for his minor corrections to this work. This work was supported by the Institute of Information \& Communications Technology Planning \& Evaluation (IITP) grant funded by the Korea government (MSIT) [RS-2021-II211341, Artificial Intelligent Graduate School Program (Chung-Ang University)]. This research was supported by the Chung-Ang University Graduate Research Scholarship in 2025.

\bibliography{main}
\clearpage
\appendix

\section{Implementation Details}
When retrieving from the datastore in all languages, we utilize the approach outlined in \cite{chirkova-etal-2024-retrieval} as our baseline. Specifically, we employ the basic\_translated\_langspec prompt template, as detailed in Table~\ref{tab:system_prompts} to generate our final mRAG answer from the generator. In our method, we retrieve and re-rank the top-50 documents for each query, and then use only the top-5 documents to generate the final answer. The document retrieval and re-ranking are carried out using bge-m3. We do not translate documents already in query language in the framework of DKM-RAG to reduce costs.

We conduct our experiments using an AMD EPYC 7313 CPU (3.0 GHz) paired with four NVIDIA RTX 4090 GPUs. We use Python 3.11.5 and PyTorch 2.3.1 for the software environment.
\begin{table}[ht]
\centering
\small
\renewcommand{\arraystretch}{1.2}
\setlength{\tabcolsep}{8pt}
\begin{tabular}{lcc}
\toprule
\textbf{Language} & \textbf{Passage Count (M)} & \textbf{Percentage (\%)} \\
\midrule
ja & 27   & 20.53 \\
en & 25   & 19.00 \\
de & 14   & 10.64 \\
fr & 13   & 9.88  \\
zh & 11   & 8.36  \\
es & 10   & 7.60  \\
ru & 8.6  & 6.54  \\
it & 8.2  & 6.23  \\
pt & 4.7  & 3.57  \\
th & 3.7  & 2.81  \\
ar & 3.3  & 2.51  \\
ko & 1.6  & 1.22  \\
fi & 1.5  & 1.14  \\
\bottomrule
\end{tabular}
\caption{Language distribution of wikipedia we use in our experiment.}
\label{tab:wiki_ratio}
\end{table}

\section{Prompts}
\label{appendix:prompts}
As shown in Table~\ref{tab:system_prompts}, we provide the prompts used to generate our final answer with the retrieved documents in our mRAG baseline. \textit{Docs} refers to retrieved documents and question refers to the current query. We also provide prompts during the passage rewriting phase in the DKM-RAG framework as stated in Table~\ref{tab:prompt_dkm}. We only provide english prompts for simplicity. And we provide prompts to measure the language preference of GPT-4o-mini, regarding answering in the specific languages as stated in Table~\ref{tab:prompt_generation}.

\begin{table*}[ht]
\centering
\renewcommand{\arraystretch}{1.3}
\setlength{\tabcolsep}{8pt}
\begin{tabular}{ll}
\toprule
\textbf{System}            & \textbf{Prompt} \\ 
\midrule
With Documents             & 
\begin{tabular}[t]{@{}l@{}} 
\texttt{You are a helpful assistant. Your task is to extract relevant}\\ 
\texttt{information from provided documents and to answer to}\\ 
\texttt{questions as short as possible. Please reply in English.} \\ 
\texttt{user: f"Background:\{docs\}\textbackslash n\textbackslash nQuestion:\{question\}"}
\end{tabular} \\ 
\midrule
Without Documents          & 
\begin{tabular}[t]{@{}l@{}} 
\texttt{You are a helpful assistant. Answer the questions as short}\\ 
\texttt{as possible. Please reply in English.}\\ 
\texttt{user: f"Question:\{question\}"}
\end{tabular} \\
\bottomrule
\end{tabular}
\caption{System prompts with and without documents. The table outlines how instructions and prompts differ when documents are provided or omitted.}
\label{tab:system_prompts}
\end{table*}
\begin{table*}[ht]
\centering
\renewcommand{\arraystretch}{1.3}
\setlength{\tabcolsep}{6pt}
\begin{tabular}{p{0.9\linewidth}}
\toprule
\textbf{Prompt} \\
\midrule
Original Passage: \{passage\} \\[1ex]
Question: \{question\} \\[1ex]
Please create an independent document according to the following requirements: \\[1ex]
1) Utilize known facts (parametric knowledge) related to the question. \\[1ex]
2) Seamlessly combine with the original passage by removing redundant or unnecessary sentences. No additional explanations are allowed. \\[1ex]
3) All content must be written smoothly and concisely in English. \\
\bottomrule

\end{tabular}
\caption{The prompt used for generating $P_\text{refined}$ based on the passage and question. The instructions guide the generator to combine parametric knowledge with the original passage while ensuring clarity and conciseness.}
\label{tab:prompt_dkm}
\end{table*}

\begin{table*}[ht]
\centering
\begin{tabular}{|p{3cm}|p{13cm}|}
\hline
 & \textbf{Content} \\
\hline
\textbf{System Message} &
\begin{minipage}[t]{\linewidth}
\texttt{You are a highly capable multilingual assistant. \\ Here are some reference documents:}
\texttt{\ \ \ \ \{top5\_passages\}}\\[0.5em]
\texttt{The user wants answers in multiple languages. \\ Please follow these rules strictly:}\\[0.5em]
\texttt{1) Return your final answer as a valid JSON object.}\\[0.5em]
\texttt{2) The JSON object must contain exactly these keys: \{TARGET\_LANGUAGES\}.}\\[0.5em]
\texttt{3) Each field's value must be the answer written in that respective language.}\\[0.5em]
\texttt{4) Do not include any additional text outside the JSON (e.g., no Markdown or explanations).}\\[0.5em]
\texttt{5) Ensure it is valid JSON with correct format.}
\end{minipage} \\
\hline
\textbf{User Message} &
\begin{minipage}[t]{\linewidth}
\texttt{Question: \{question\}}\\[0.5em]
\texttt{Please provide the answers in JSON form for each of the following languages: \{TARGET\_LANGUAGES\}.}
\end{minipage} \\
\hline
\end{tabular}
%\vspace{-1mm}
\caption{Prompts used for measuring language preference of GPT-4o-mini in mRAG pipeline.}
\label{tab:prompt_generation}
\end{table*}

\section{Language Notation}
In this work, we use standard ISO 639-1 language codes to represent the various languages involved in our experiments. Specifically, en denotes English, ko represents Korean, ar corresponds to Arabic, zh refers to Chinese (Simplified), fi indicates Finnish, fr stands for French, de represents German, ja corresponds to Japanese, it refers to Italian, pt denotes Portuguese, ru stands for Russian, es represents Spanish, and th corresponds to Thai. These concise notations facilitate the identification and processing of language-specific data across datasets and models in multilingual NLP research.

\section{Dataset Statistics}

We present the statistics of the datasets used in our experiments. MKQA serves as the primary dataset, and its details, including the number of examples and the median lengths of questions and answers, are summarized in Table~\ref{tab:data_statistics}. Additionally, we utilize Wikipedia as the external source for the retriever datastore, with its statistics (number of passages and median lengths) also provided in Table~\ref{tab:data_statistics}. And we provide the number of passages in each language and the ratio of them in Table~\ref{tab:wiki_ratio}. These details offer a clear overview of the data resources supporting our experiments.

\paragraph{Language Distribution of Pre-trained LLM}
We provide language distribution in the pre-training corpus of Llama-2. As stated in Table~\ref{tab:llama_distribution}, we use English (EN) as a high-resource, Spanish (ES) as a mid-resource, and Korean (KO) as a low-resource language in our experiment based on their ratios.
\begin{table*}[ht]
\centering
\begin{tabular}{l r}
\hline
\textbf{Language} & \textbf{Percentage} \\
\hline
\textbf{EN}       & \textbf{89.70}\% \\
Unknown  & 8.38\%  \\
DE       & 0.17\%  \\
FR       & 0.15\%  \\
SV       & 0.15\%  \\
\textbf{ES}       & \textbf{0.13}\%  \\
ZH       & 0.15\%  \\
RU       & 0.12\%  \\
NL       & 0.11\%  \\
IT       & 0.11\%  \\
JP       & 0.11\%  \\
PL       & 0.09\%  \\
PT       & 0.09\%  \\
VI       & 0.08\%  \\
RO       & 0.03\%  \\
SR       & 0.04\%  \\
CA       & 0.04\%  \\
\textbf{KO}       & \textbf{0.06}\%  \\
UK       & 0.07\%  \\
Other    & 0.21\%  \\
\hline
\end{tabular}
\caption{Language distribution in the pre-training corpus of Llama-2. Unknown represents languages we cannot know because of closed-source access of model and other denotes other languages.}
\label{tab:llama_distribution}
\end{table*}

\label{appendix:statistics}

\begin{table*}[ht]
\centering
\small
\renewcommand{\arraystretch}{1.2}
\setlength{\tabcolsep}{6pt}
\begin{tabular}{lccccccccccccc}
\toprule
\textbf{Dataset} & \textbf{en} & \textbf{ar} & \textbf{es} & \textbf{fi} & \textbf{fr} & \textbf{de} & \textbf{ja} & \textbf{it} & \textbf{ko} & \textbf{pt} & \textbf{ru} & \textbf{zh} & \textbf{th}\\
\midrule
\multicolumn{14}{l}{\textbf{MKQA}} \\
\# examples & 2827 & 2827 & 2827 & 2827 & 2827 & 2827 & 2827 & 2827 & 2827 & 2827 & 2827 & 2827 & 2827 \\
len question.   & 43   & 38   & 48   & 46   & 49   & 47   & 26   & 48   & 22   & 45   & 42   & 16   & 41\\
len answer.   & 11   & 10   & 11   & 11   & 11   & 11   &  8   & 11   &  6   & 11   & 12   &  6  & 12 \\
\midrule
\multicolumn{14}{l}{\textbf{Wikipedia}} \\
\# ex. (M)  & 25   & 3.3  & 10   & 1.5  & 13   & 14   & 27   & 8.2  & 1.6  & 4.7  & 8.6  & 11   & 3.7\\
len passage.   & 624  & 585  & 619  & 833  & 627  & 720  & 208  & 650  & 431  & 619  & 721  & 206  & 217\\
\bottomrule
\end{tabular}
%\vspace{-3mm}
\caption{Statistics of the datasets used in our experiments. MKQA Number of examples and median lengths of questions and answers (in Unicode characters). Wikipedia: Number of passages (in millions) and their median lengths.}
\label{tab:data_statistics}
\end{table*}

\section{Language Preference of Other Languages}
We also perform additional experiments to explore language preferences for languages not covered in Table~\ref{tab:subset_mlr}, using the MLRS score that we propose. As shown in Table~\ref{tab:appendix_removed_langs}, similar to the results in Table~\ref{tab:subset_mlr}, the highest preferences are typically observed when $L_q = L_d$ across all query languages. English is also the most preferred language. For clarity, we omit results for other languages.

For most languages, such as Arabic, Finnish, German, and Russian, switching to a cross-lingual setup leads to a significant drop in MLRS. For example, Arabic queries using the bge-m3 encoder achieve a monolingual score of 40.39, but cross-lingual retrieval (e.g., with Thai) results in a 6.80-point decrease.

Interestingly, for Thai queries, some cross-lingual pairs show a slight improvement over the monolingual baseline (as indicated by the positive differences in red), suggesting that for low-resource languages like Thai, cross-lingual signals might sometimes offer complementary benefits

\begin{table*}[!ht]
\centering
\setlength{\tabcolsep}{3pt}
\renewcommand{\arraystretch}{1.1}
\resizebox{\textwidth}{!}{
\begin{tabular}{cc|>{\columncolor{gray!15}}c|ccccc}
\toprule
\multicolumn{2}{c|}{} 
& \textbf{\(L_q=L_d\)} 
& \multicolumn{5}{c}{\textbf{\(L_q \neq L_d\)}} \\
\textbf{Query Lang.} & \textbf{Encoder} 
& 
& \textbf{ar} & \textbf{fi} & \textbf{de} & \textbf{ru} & \textbf{th} \\
\midrule
% =========================
% QL = ar
% =========================
\multirow{3}{*}{\textbf{ar}}
& \textbf{bge-m3}
  & \textbf{40.39} 
  & -- 
  & 34.10 {\scriptsize(\textcolor{blue}{-6.29})}
  & 35.91 {\scriptsize(\textcolor{blue}{-4.48})}
  & \underline{36.22} {\scriptsize(\textcolor{blue}{-4.17})}
  & 33.59 {\scriptsize(\textcolor{blue}{-6.80})} \\
& \textbf{p-mMiniLM}
  & \textbf{41.25}
  & --
  & 34.90 {\scriptsize(\textcolor{blue}{-6.35})}
  & 36.58 {\scriptsize(\textcolor{blue}{-4.67})}
  & \underline{37.13} {\scriptsize(\textcolor{blue}{-4.12})}
  & 34.46 {\scriptsize(\textcolor{blue}{-6.79})} \\
& \textbf{p-mMpNet}
  & \textbf{41.34}
  & --
  & 34.64 {\scriptsize(\textcolor{blue}{-6.70})}
  & 36.34 {\scriptsize(\textcolor{blue}{-5.00})}
  & \underline{36.87} {\scriptsize(\textcolor{blue}{-4.47})}
  & 34.36 {\scriptsize(\textcolor{blue}{-6.98})} \\
\midrule
% =========================
% QL = fi
% =========================
\multirow{3}{*}{\textbf{fi}}
& \textbf{bge-m3}
  & \textbf{36.65}
  & 33.47 {\scriptsize(\textcolor{blue}{-3.18})}
  & --
  & \underline{36.33} {\scriptsize(\textcolor{blue}{-0.32})}
  & 35.42 {\scriptsize(\textcolor{blue}{-1.23})}
  & 33.07 {\scriptsize(\textcolor{blue}{-3.58})} \\
& \textbf{p-mMiniLM}
  & \textbf{37.37}
  & 34.60 {\scriptsize(\textcolor{blue}{-2.77})}
  & --
  & \underline{37.14} {\scriptsize(\textcolor{blue}{-0.23})}
  & 36.48 {\scriptsize(\textcolor{blue}{-0.89})}
  & 34.12 {\scriptsize(\textcolor{blue}{-3.25})} \\
& \textbf{p-mMpNet}
  & \textbf{37.27}
  & 34.41 {\scriptsize(\textcolor{blue}{-2.86})}
  & --
  & \underline{36.92} {\scriptsize(\textcolor{blue}{-0.35})}
  & 36.28 {\scriptsize(\textcolor{blue}{-0.99})}
  & 34.12 {\scriptsize(\textcolor{blue}{-3.15})} \\
\midrule
% =========================
% QL = de
% =========================
\multirow{3}{*}{\textbf{de}}
& \textbf{bge-m3}
  & \textbf{39.81}
  & 33.21 {\scriptsize(\textcolor{blue}{-6.60})}
  & 34.16 {\scriptsize(\textcolor{blue}{-5.65})}
  & --
  & \underline{34.63} {\scriptsize(\textcolor{blue}{-5.18})}
  & 32.95 {\scriptsize(\textcolor{blue}{-6.86})} \\
& \textbf{p-mMiniLM}
  & \textbf{40.80}
  & 34.62 {\scriptsize(\textcolor{blue}{-6.18})}
  & 35.25 {\scriptsize(\textcolor{blue}{-5.55})}
  & --
  & \underline{35.94} {\scriptsize(\textcolor{blue}{-4.86})}
  & 34.18 {\scriptsize(\textcolor{blue}{-6.62})} \\
& \textbf{p-mMpNet}
  & \textbf{40.92}
  & 34.81 {\scriptsize(\textcolor{blue}{-6.11})}
  & 35.33 {\scriptsize(\textcolor{blue}{-5.59})}
  & --
  & \underline{36.13} {\scriptsize(\textcolor{blue}{-4.79})}
  & 34.37 {\scriptsize(\textcolor{blue}{-6.55})} \\
\midrule
% =========================
% QL = ru
% =========================
\multirow{3}{*}{\textbf{ru}}
& \textbf{bge-m3}
  & \textbf{45.05}
  & 33.84 {\scriptsize(\textcolor{blue}{-11.21})}
  & 34.20 {\scriptsize(\textcolor{blue}{-10.85})}
  & \underline{35.63} {\scriptsize(\textcolor{blue}{-9.42})}
  & --
  & 33.24 {\scriptsize(\textcolor{blue}{-11.81})} \\
& \textbf{p-mMiniLM}
  & \textbf{46.08}
  & 34.85 {\scriptsize(\textcolor{blue}{-11.23})}
  & 35.18 {\scriptsize(\textcolor{blue}{-10.90})}
  & \underline{36.73} {\scriptsize(\textcolor{blue}{-9.35})}
  & --
  & 34.23 {\scriptsize(\textcolor{blue}{-11.85})} \\
& \textbf{p-mMpNet}
  & \textbf{45.82}
  & 34.63 {\scriptsize(\textcolor{blue}{-11.19})}
  & 34.83 {\scriptsize(\textcolor{blue}{-10.99})}
  & \underline{36.28} {\scriptsize(\textcolor{blue}{-9.54})}
  & --
  & 34.12 {\scriptsize(\textcolor{blue}{-11.70})} \\
\midrule
% =========================
% QL = th
% =========================
\multirow{3}{*}{\textbf{th}}
& \textbf{bge-m3}
  & 34.52
  & 33.68 {\scriptsize(\textcolor{blue}{-0.84})}
  & 34.11 {\scriptsize(\textcolor{blue}{-0.41})}
  & \textbf{35.99} {\scriptsize(\textcolor{red}{+1.47})}
  & \underline{35.60} {\scriptsize(\textcolor{red}{+1.08})}
  & -- \\
& \textbf{p-mMiniLM}
  & 35.38
  & 34.65 {\scriptsize(\textcolor{blue}{-0.73})}
  & 34.77 {\scriptsize(\textcolor{blue}{-0.61})}
  & \textbf{36.63} {\scriptsize(\textcolor{red}{+1.25})}
  & \underline{36.40} {\scriptsize(\textcolor{red}{+1.02})}
  & -- \\
& \textbf{p-mMpNet}
  & 34.73
  & 34.10 {\scriptsize(\textcolor{blue}{-0.63})}
  & 34.14 {\scriptsize(\textcolor{blue}{-0.59})}
  & \textbf{36.08} {\scriptsize(\textcolor{red}{+1.35})}
  & \underline{35.84} {\scriptsize(\textcolor{red}{+1.11})}
  & -- \\
\bottomrule
\end{tabular}
}
\caption{
Language preference measured by MLRS with various re-ranking encoders for various query and document language combinations in a multilingual retriever. The \(L_q=L_d\) column reports the diagonal scores where the query language matches the translated document language, while the remaining columns represent cross-lingual scenarios (i.e., where the query language differs from the document language). Scores in parentheses indicate the difference from the diagonal value (\textcolor{red}{positive} for an improvement, \textcolor{blue}{negative} for a decline). The highest score for each row is highlighted in bold, and the second highest is underlined.
}
\label{tab:appendix_removed_langs}
\end{table*}

\section{Similarity Matrices}
\begin{figure*}[ht]
  \centering
  \includegraphics[width=1.0\textwidth]{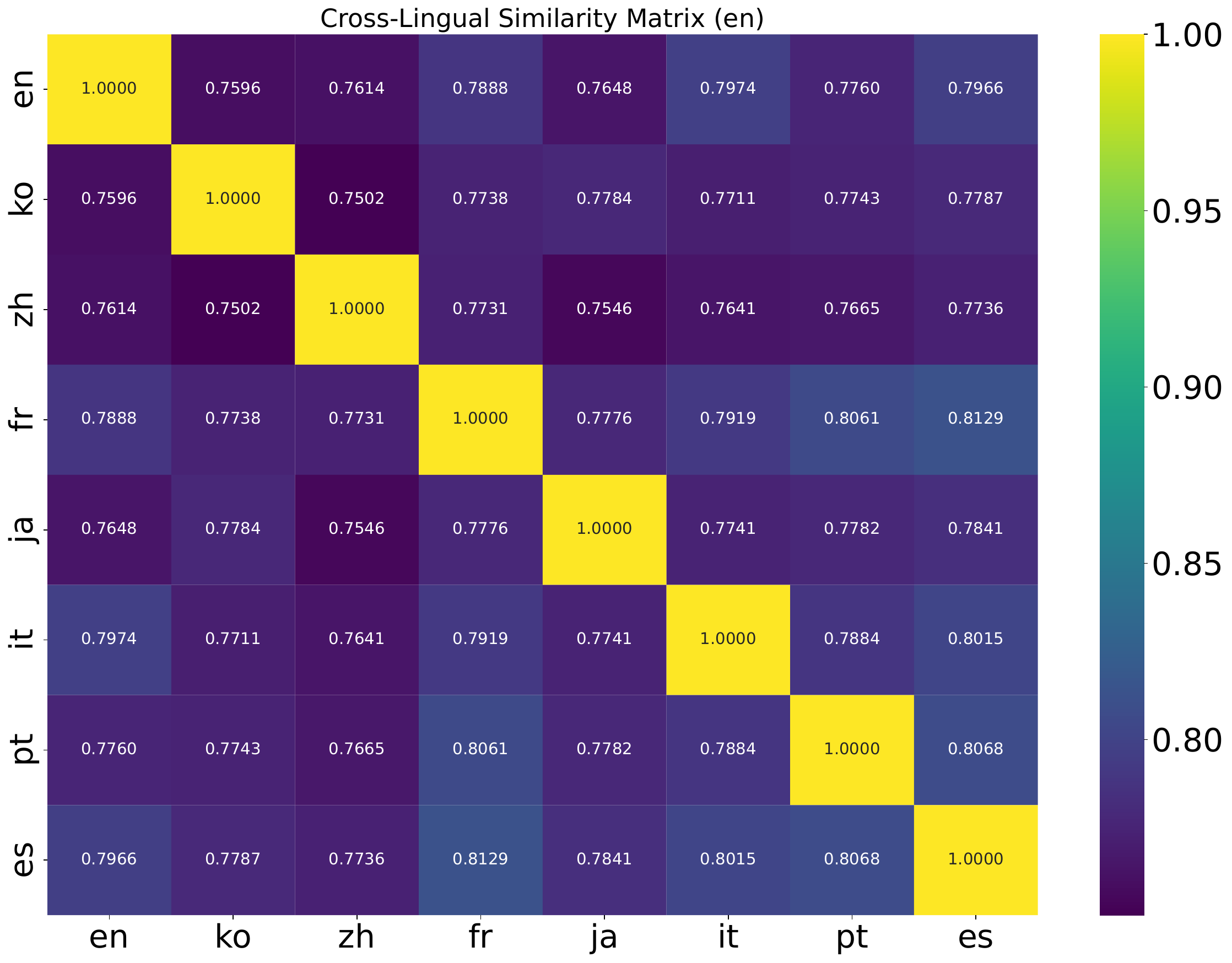}
  \caption{LaBSE Similarity Matrix of aya-expanse-8b (en).} 
  \label{fig:aya_lang_pref_en}
\end{figure*}

\begin{figure*}[ht]
  \centering
  \includegraphics[width=1.0\textwidth]{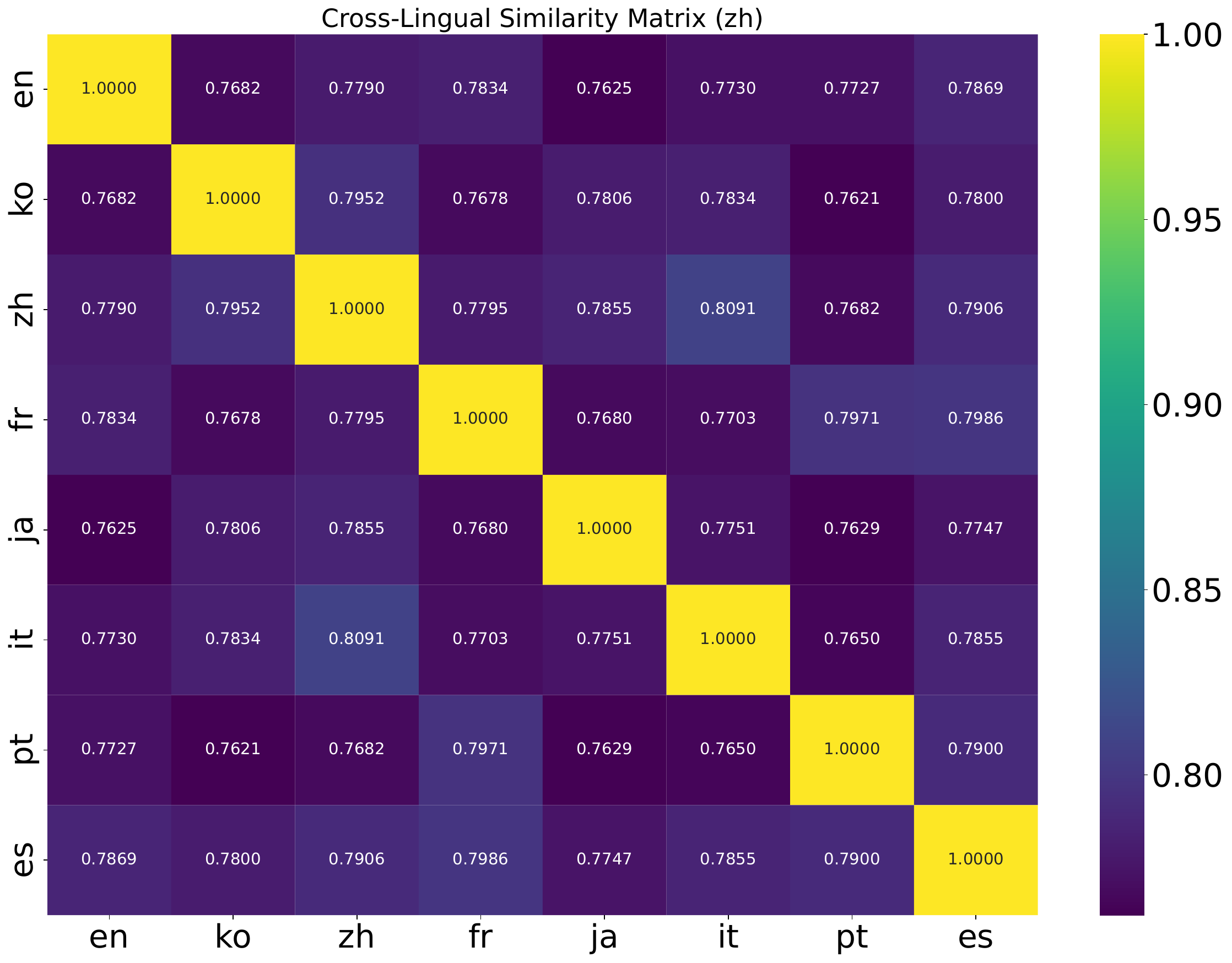} 
  \caption{LaBSE Similarity Matrix (zh) of aya-expanse-8b.} 
  \label{fig:aya_lang_pref_zh}
\end{figure*}

\begin{figure*}[ht]
  \centering
  \includegraphics[width=1.0\textwidth]{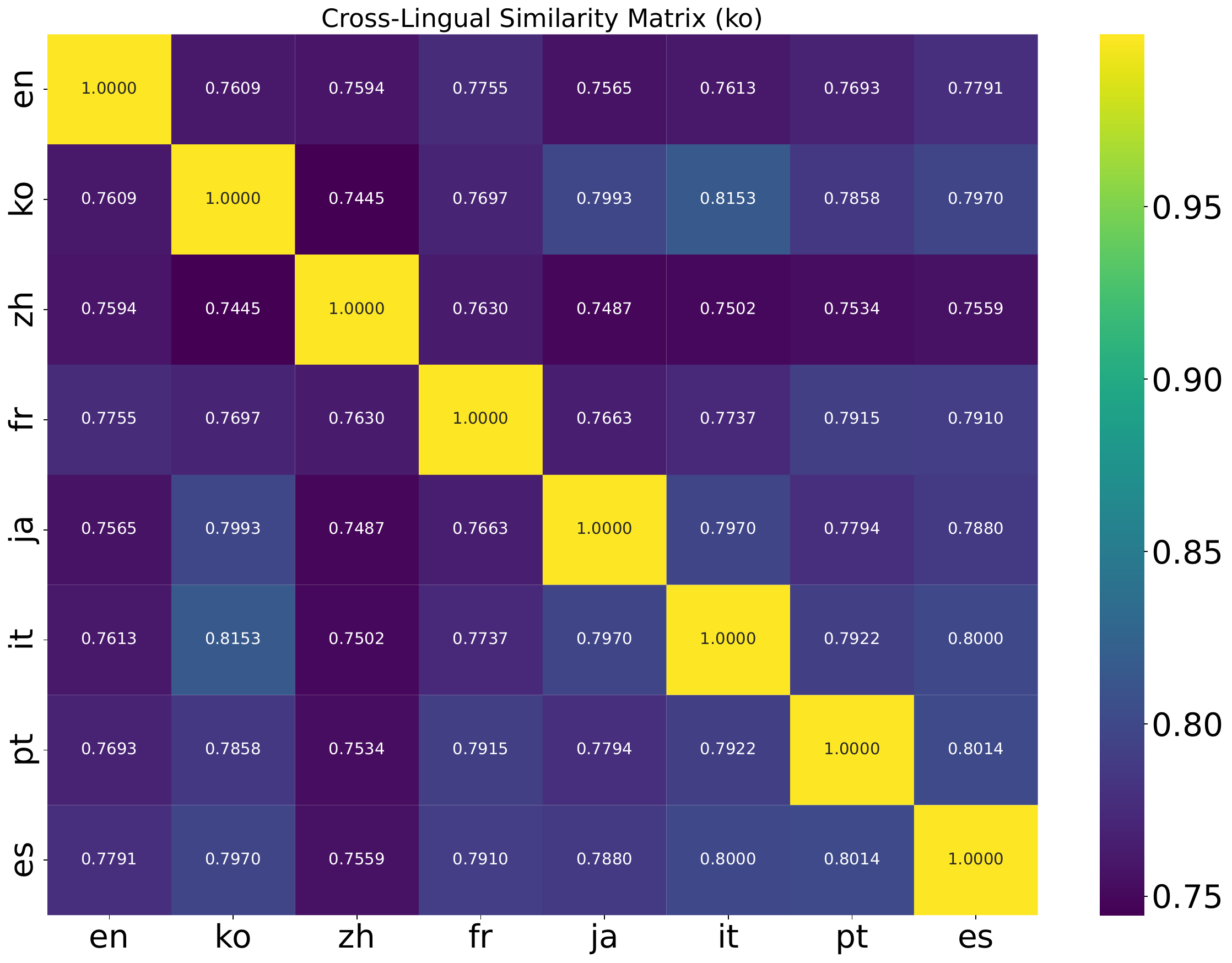}
  \caption{LaBSE Similarity Matrix (ko) of aya-expanse-8b.} 
  \label{fig:aya_lang_pref_ko}
\end{figure*}

\begin{figure*}[ht]
  \centering
  \includegraphics[width=1.0\textwidth]{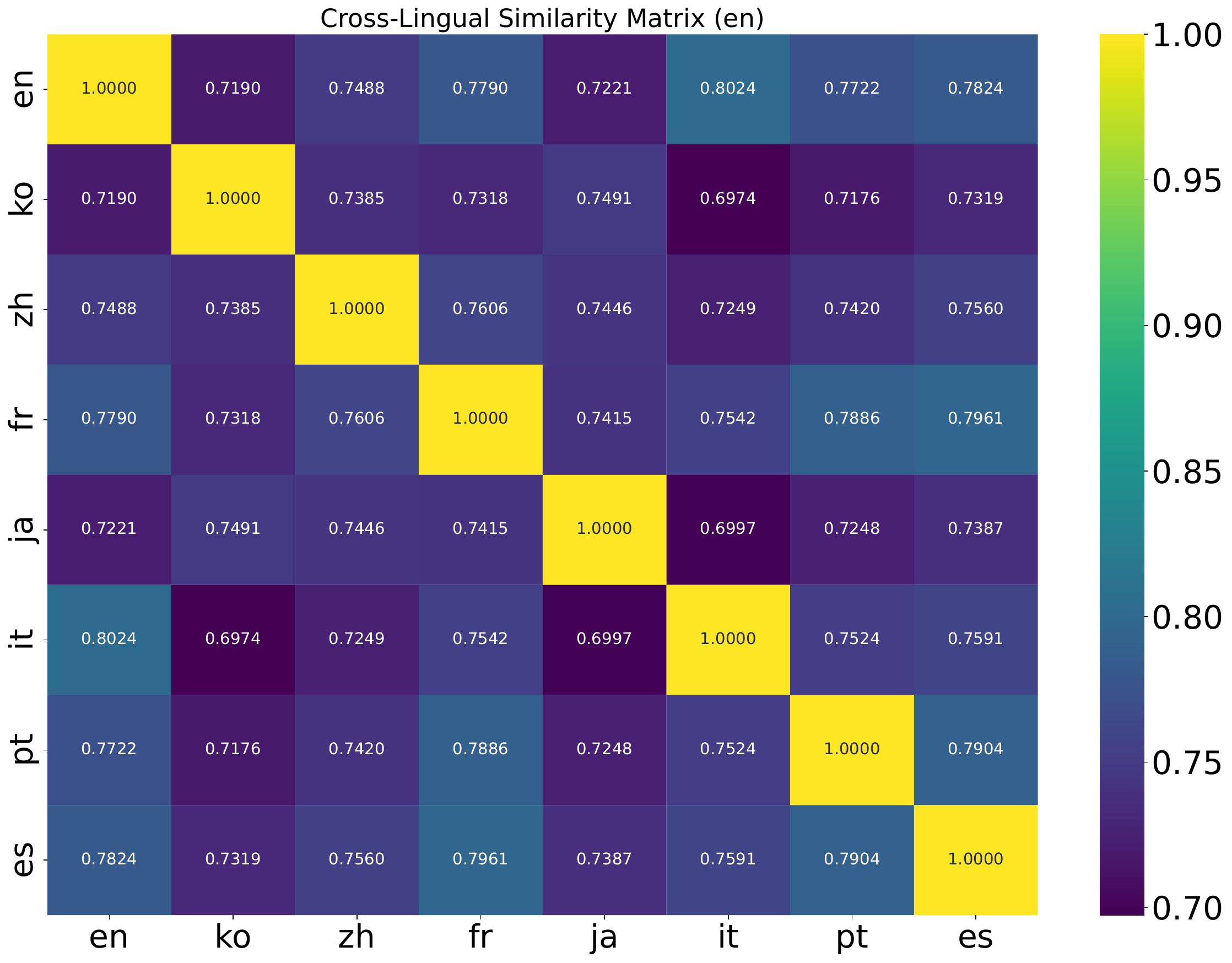}
  \caption{LaBSE Similarity Matrix (en) of Llama-3.1-8B-instruct.} 
  \label{fig:llama_lang_pref_en}
\end{figure*}

\begin{figure*}[ht]
  \centering
  \includegraphics[width=1.0\textwidth]{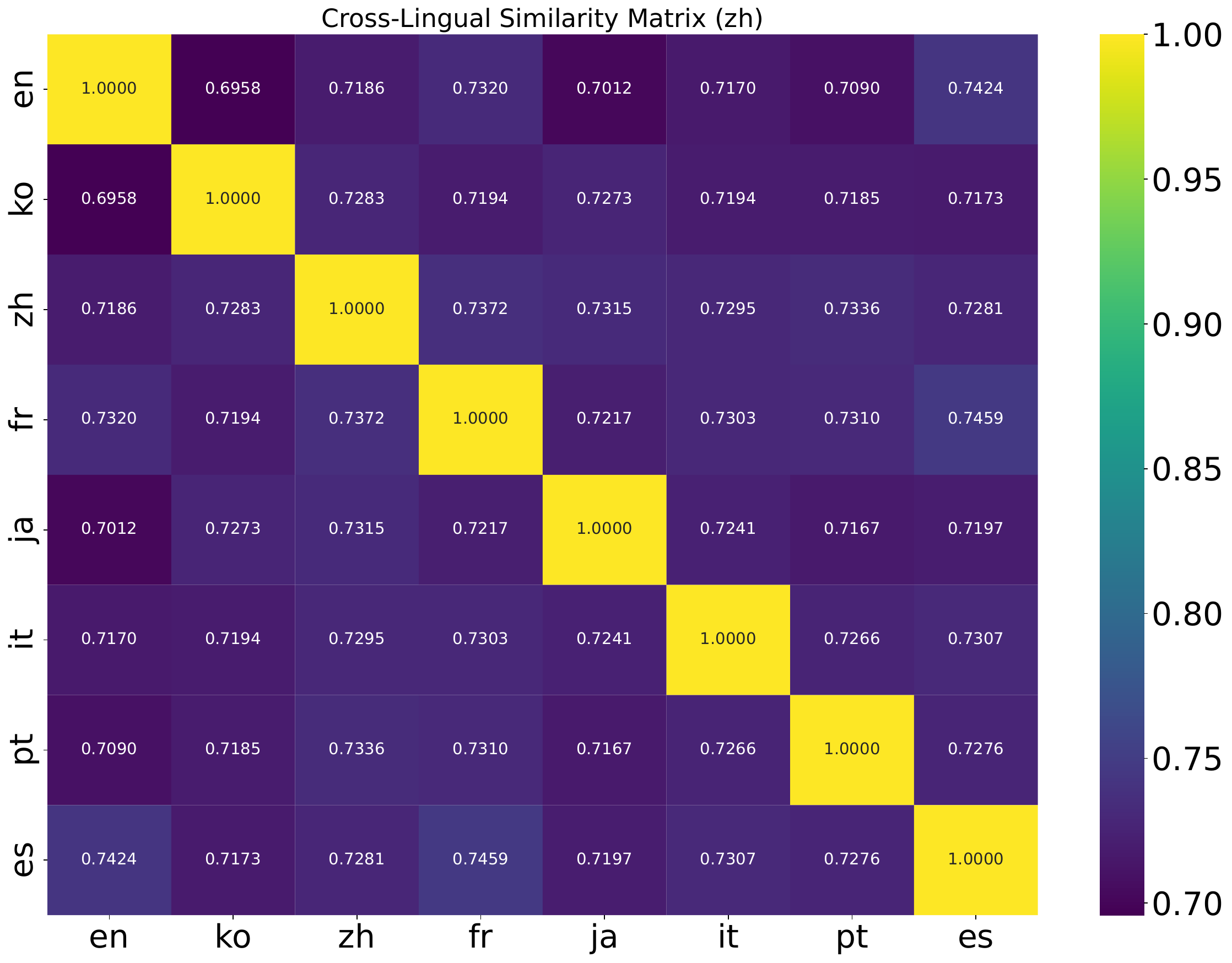} 
  \caption{LaBSE Similarity Matrix (zh) of Llama-3.1-8B-instruct.} 
  \label{fig:llama_lang_pref_zh}
\end{figure*}

\begin{figure*}[ht]
  \centering
  \includegraphics[width=1.0\textwidth]{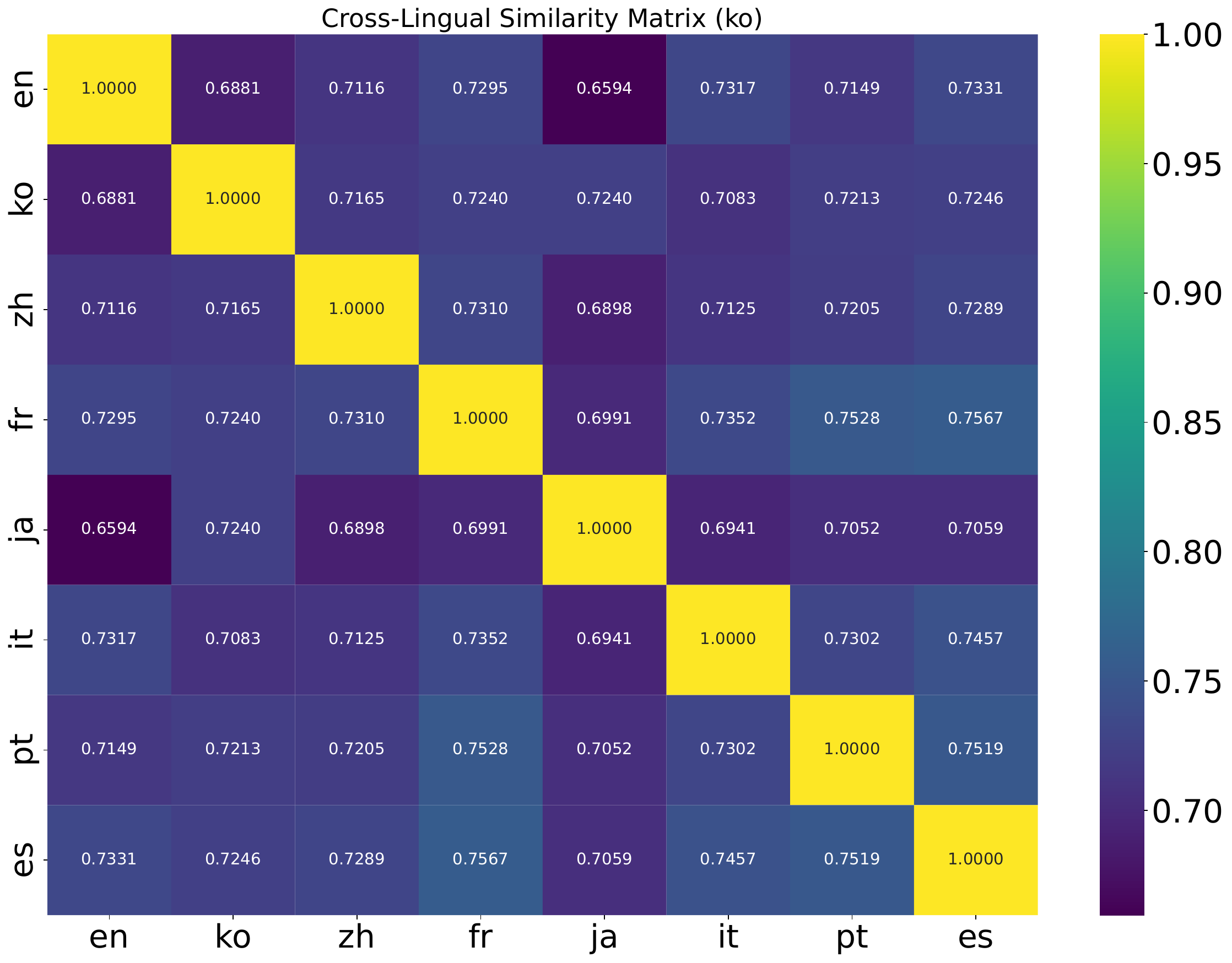}
  \caption{LaBSE Similarity Matrix (ko) of Llama-3.1-8B-instruct.} 
  \label{fig:llama_lang_pref_ko}
\end{figure*}

\begin{figure*}[ht]
  \centering
  \includegraphics[width=1.0\textwidth]{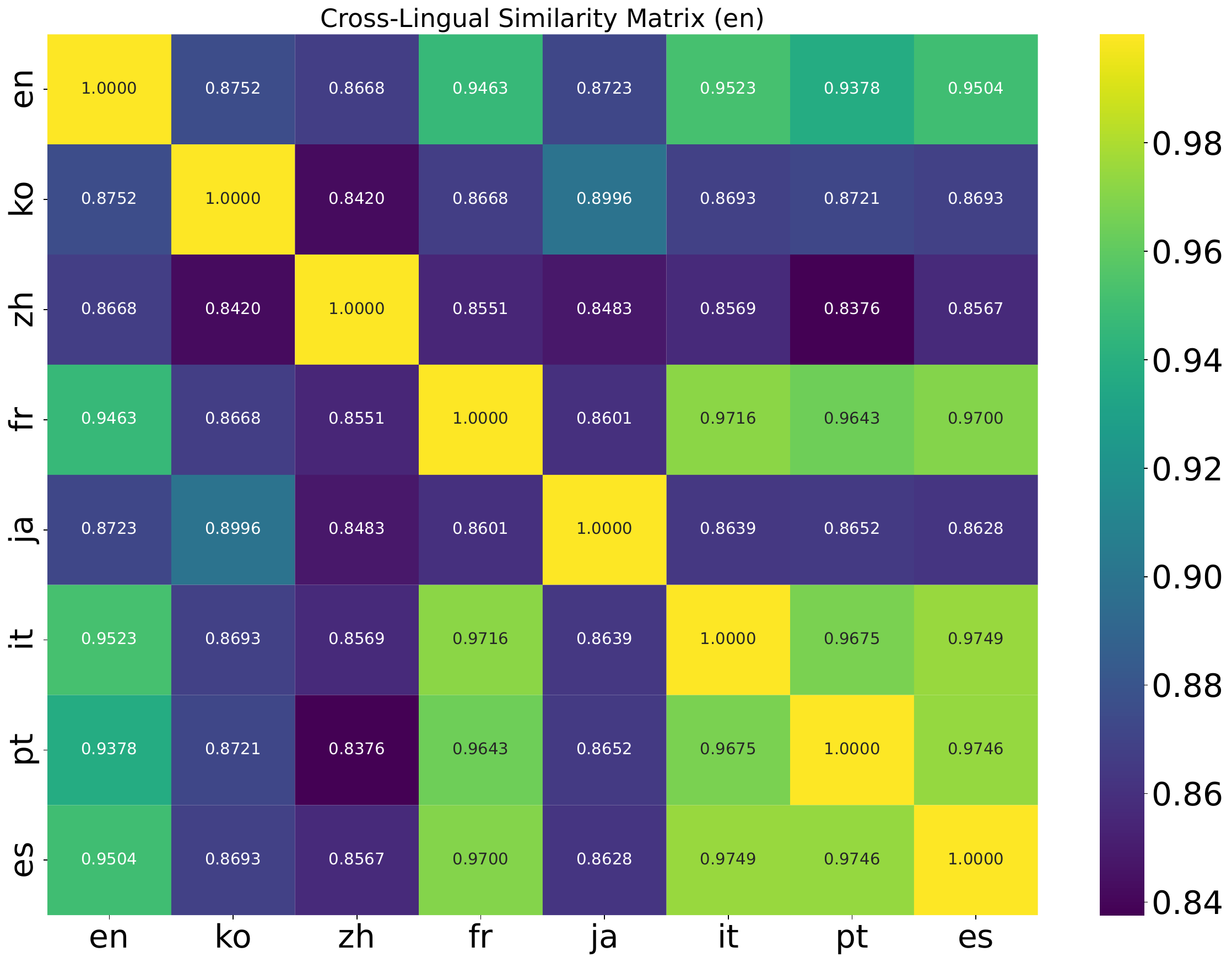}
  \caption{LaBSE Similarity Matrix (en) of gpt-4o-mini.} 
  \label{fig:gpt_lang_pref_en}
\end{figure*}

\begin{figure*}[ht]
  \centering
  \includegraphics[width=1.0\textwidth]{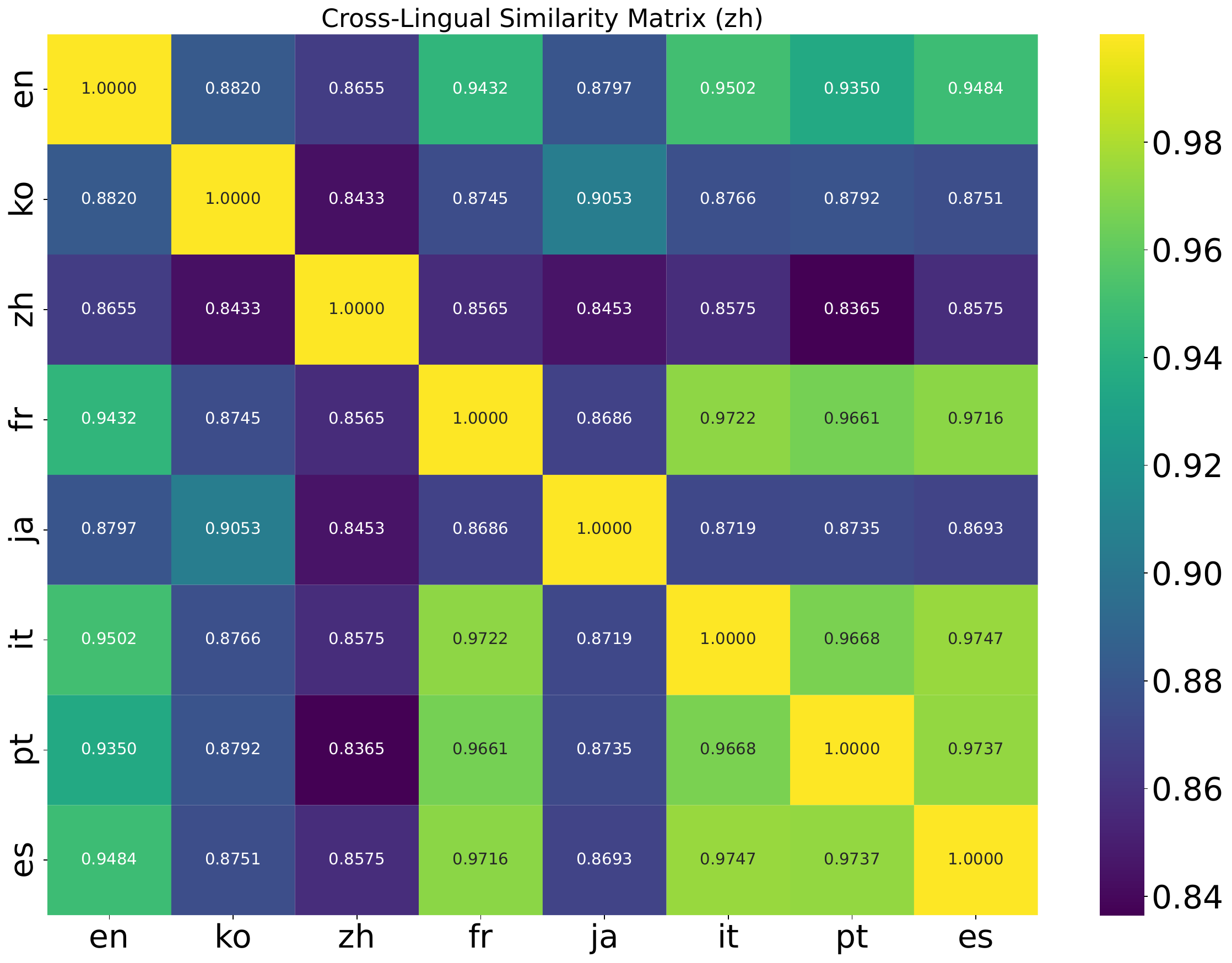} 
  \caption{LaBSE Similarity Matrix (zh) of gpt-4o-mini.} 
  \label{fig:gpt_lang_pref_zh}
\end{figure*}

\begin{figure*}[ht]
  \centering
  \includegraphics[width=1.0\textwidth]{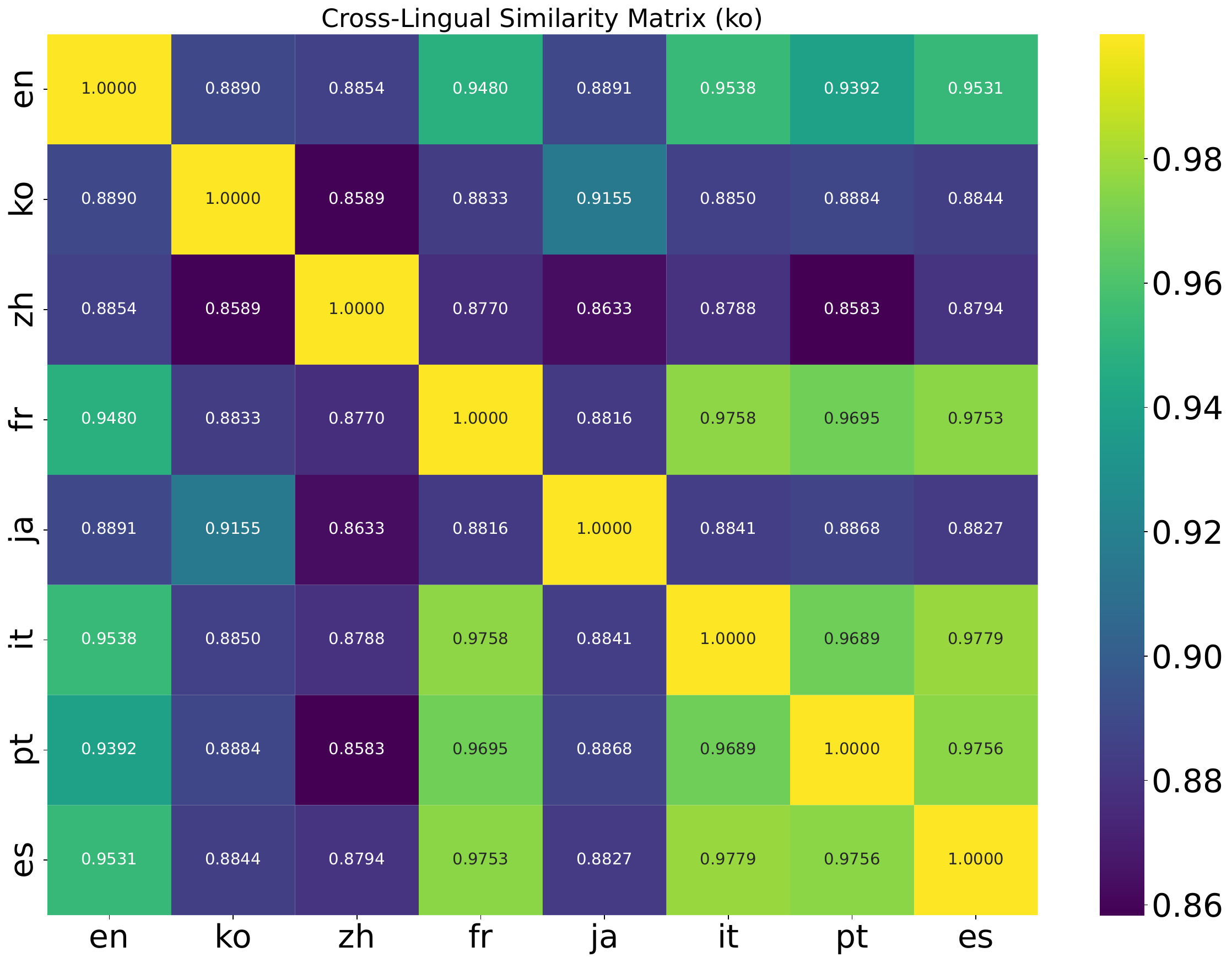}
  \caption{LaBSE Similarity Matrix (ko) of gpt-4o-mini.} 
  \label{fig:gpt_lang_pref_ko}
\end{figure*}

We provide similarity matrix measured by LaBSE for each query language en, zh, ko and each generator in Figure~\ref{fig:aya_lang_pref_en}, Figure~\ref{fig:aya_lang_pref_zh}, Figure~\ref{fig:aya_lang_pref_ko}, Figure~\ref{fig:llama_lang_pref_en}, Figure~\ref{fig:llama_lang_pref_zh}, Figure~\ref{fig:llama_lang_pref_ko}, Figure~\ref{fig:gpt_lang_pref_en}, Figure~\ref{fig:gpt_lang_pref_zh} and Figure~\ref{fig:gpt_lang_pref_ko}. Each entry represents the embedding similarity score between answers generated in different languages, with the diagonal values all equal to 1 (i.e., comparing an answer with itself). Moreover, the values shown in Figure~\ref{fig:multi_lang} are computed by averaging over the rows or columns for each language.

\section{Case study}
\paragraph{MLRS}
We provide an example of a document that improved MLRS score, where the rank of a relevant document significantly increases after translation. In Table~\ref{tab:case_mlr}, the user query \textit{"영국 캐리비안에 언제 노예제가 폐지됐나요? (When was slavery abolished in the British Caribbean?)"} is in Korean, whereas the original passage is in English. Initially, the document’s rank (\(\mathbf{r_d^{\text{init}}} = 34\)) was relatively low, but after translating the passage into Korean and re-ranking (\(\mathbf{r_d^{\text{re-rank}}} = 2\)), the document moved much closer to the top. 
This demonstrates how cross-lingual alignment can substantially improve retrieval performance in a multilingual setting.
Notably, even if the passage content is semantically the same, language preference in the model can lead to poor alignment when the query and document are in different languages, adversely affecting retrieval. Translating the document into the query language effectively mitigates this issue.
\begin{table*}[ht]
\centering
\begin{tabularx}{\textwidth}{lX}
\hline
\textbf{Field} & \textbf{Value} \\
\hline
\textbf{query} & 영국 캐리비안에 언제 노예제가 폐지됐나요? (When was slavery abolished in the British Caribbean?) \\
\hline
\textbf{gold answer} & \textcolor{red}{1834-08-01} \\
\hline
\textbf{doc id} & kilt-100w\_6947054 (English) \\
\hline
\(\mathbf{r_d^{\text{init}}}\) & 34 \\
\hline
\(\mathbf{r_d^{\text{re-rank}}}\) & 2 \\
\hline
\textbf{d (content)} & \textit{History of the Caribbean. Empire remained slaves, however, until Britain passed the Slavery Abolition Act in 1833. 
When the Slavery Abolition Act came into force in 1834, roughly 700,000 slaves in the British West Indies immediately became free; 
other enslaved workers were freed several years later after a period of forced apprenticeship. 
Slavery was abolished in the Dutch Empire in 1814. 
Spain abolished slavery in its empire in 1811, with the exceptions of Cuba, Puerto Rico, and Santo Domingo; 
Spain ended the slave trade to these colonies in 1817, after being paid ₤400,000 by Britain. 
Slavery itself was not abolished in Cuba until 1886.} \\
\hline
\textbf{d (translated)} & \textit{1834년 노예제 폐지법이 시행되자, 영국 서인도 제도에서 약 700,000명의 노예가 즉시 해방되었고, 
다른 노예 노동자들은 강제 연습생 생활을 한 후 몇 년 후에 해방되었다. 
1814년 네덜란드 제국에서 노예제는 폐지되었다. 
1811년 스페인은 쿠바, 푸에르토리코, 산토 도밍고를 제외하고는 제국에서 노예제를 폐지했다. 
1817년 영국이 400만원을 지불한 후 스페인은 이들 식민지에서의 노예 무역을 종식시켰다. 
노예제는 1886년까지 쿠바에서 폐지되지 않았다.} \\
\hline
\end{tabularx}
\caption{An example of an improved MLRS case. After translating the document into Korean, its rank improved from 34 to 2, illustrating language preference of retriever.}
\label{tab:case_mlr}
\end{table*}

\paragraph{Answer Generation in Language Preference of Generator}

\begin{table*}[ht]
\centering
% Metadata 부분
\begin{tabular}{ll}
\toprule
\textbf{Question} & which type of air pressure is associated with warm air rising \\
\bottomrule
\end{tabular}

%\vspace{1em}

\begin{tabular}{l p{10cm} c}
\toprule
\textbf{Language} & \textbf{Answer} & \textbf{Preference Score} \\
\midrule
en & Low pressure is associated with warm air rising. & 0.9179 \\
ko & 따뜻한 공기가 상승하는 것과 관련된 공기 압력은 저압입니다. & 0.9060 \\
zh & 与暖空气上升相关的空气压力是低压。 & 0.8853 \\
fr & La pression basse est associée à l'ascension de l'air chaud. & 0.9231 \\
ja & 暖かい空気が上昇することに関連する気圧は低圧です。 & 0.9187 \\
it & La bassa pressione è associata all'aria calda che sale. & 0.9256 \\
pt & A pressão baixa está associada ao ar quente que sobe. & 0.9316 \\
es & La presión baja está asociada con el aire caliente que asciende. & 0.9317 \\
\bottomrule
\end{tabular}
\caption{An example of generated answers in different languages with gpt-4o-mini. Also, we report the average similarity score between each pair of answers.}
\label{tab:case_generation}
\end{table*}

We also provide an example of generated answers in different languages with a generator, GPT-4o-mini as shown in Table~\ref{tab:case_generation}. The preference score in the rightmost column of Table~\ref{tab:case_generation} indicates that the generator prefers the query language and Latin-script languages over other languages.

\paragraph{Unified Document of DKM-RAG}
\begin{table*}[ht]
\centering
\begin{tabularx}{\textwidth}{lX}
\hline
\textbf{Field} & \textbf{Value}\\
\hline
\(\displaystyle \text{query}\) & \textit{in the united states the president is the head of which branch of government?} \\
\hline
\(\displaystyle \text{gold answer}\) 
& \textcolor{red}{the executive branch} \\
\hline
\(\displaystyle \text{doc id}\) 
& kilt-100w\_5089743 (English)\\
\hline
\(\displaystyle P_{\text{translated}}\) 
& \textit{President of the United States. President of the United States (POTUS) 
is the head of state and head of government of the United States of America. 
The president is the commander-in-chief of the United States Armed Forces. 
In contemporary times, the president is looked upon as one of the world's 
most powerful political figures as the leader of the only remaining global superpower. The role includes responsibility for the world's most expensive military, which has the second largest nuclear arsenal. The president also leads the nation with the largest economy. } \\
\hline
\(\displaystyle P_{\text{refined}}\) 
& \textit{The president of the United States is the head of the 
\textcolor{red}{executive branch} of the federal government. 
The president directs the \textcolor{red}{executive branch} 
and is the commander-in-chief of the United States Armed Forces/ In contemporary times, the president is looked upon as one of the world's most powerful political figures as the leader of the only remaining global superpower. The role includes responsibility for the world's most expensive military, which has the second-largest nuclear arsenal. The president also leads the nation with the largest economy.} \\
\hline
\end{tabularx}
\caption{A DKM-RAG case study illustrating how \(\displaystyle P_{\text{translated}}\) and \(\displaystyle P_{\text{refined}}\) 
correspond to the retrieved passage (translated into the query language) and the rewritten passage leveraging parametric knowledge, respectively. The overlap with the gold answer is highlighted in \textcolor{red}{red}.}
\label{tab:case_dkm}
\end{table*}

Additionally, we provide a sample of \(\displaystyle P_{\text{translated}}\) and \(\displaystyle P_{\text{refined}}\) obtained via our proposed DKM-RAG framework in Table~\ref{tab:case_dkm}. 
This example illustrates how the crucial answer component, \textit{``the executive branch''}, which is not apparent from the translated passage alone, emerges through the model’s internal knowledge. 
Consequently, this shows that DKM-RAG can effectively leverage additional knowledge sources that is not included in the translated passage to achieve better performance.

%\vspace{-1mm}
\subsection{Failure Case}
\paragraph{MLRS}
We present a failure case of the MLRS metric in Table~\ref{tab:failure_mlr}. Due to the difficulty of translating documents in low-resource languages, repetitive phrases such as \textit{Changing the line-up} appear in the translated passage. This repetition causes the re-ranker to misinterpret the content, leading to an improvement in the rank even though the content is irrelevant.
\begin{table*}[ht]
\centering
\begin{tabularx}{\textwidth}{lX}
\hline
\textbf{Field} & \textbf{Value} \\
\hline
\textbf{query} & \textit{연속으로 가장 많은 자유투 기록 (who holds the record for most free throws made in a row)} \\
\hline
\textbf{gold answer} & \textcolor{red}{톰 앰베리} \\
\hline
\textbf{doc id} & wiki-100w-ja\_8993041 \\
\hline
\(\mathbf{r_d^{\text{init}}}\) & 31 \\
\hline
\(\mathbf{r_d^{\text{re-rank}}}\) & 5 \\
\hline
\textbf{d (content)} & \textit{'林直明. を 変 更 \textbackslash n \textbackslash n 記 録 \textbackslash n \quad イ ニ ン グ 最 多 連 続 与 四 球 ： 5 （ 日 本 記 録 ） \quad 1 9 4 6 年 4 月 2 9 日 \textbackslash n \quad 同 一 年 に 2 球 団 で 勝 利 ： 1 9 4 8 年 \quad ※ 史 上 3 人 目 \textbackslash n \quad ゲ ー ム 最 多 失 点 ： 1 4 （ セ ・ リ ー グ 記 録 ） \quad 1 9 5 0 年 6 月 7 日 \textbackslash n \textbackslash n 背 番 号'} \\
\hline
\textbf{d (translated)} & \textit{Changing the line-up, Changing the line-up, Changing the line-up, Changing the line-up, Changing the line-up, Changing the line-up, Changing the line-up, Changing the line-up, Changing the line-up, Changing the line-up, Changing the line-up, Changing the line-up, Changing the line-up, Changing the line-up, Changing the line-up, (...)} \\
\hline
\end{tabularx}
\caption{A failure case of MLRS because bad translation quality due to difficulty in translating low-resource language.}
\label{tab:failure_mlr}
\end{table*}

\paragraph{DKM-RAG}
We also present a failure case of DKM-RAG in Table~\ref{tab:failure_dkmrag}. The retriever retrieves an English document that is irrelevant to the query due to its language preference. Additionally, the LLM lacks relevant knowledge related to the query, resulting in a failed generation.
\begin{table*}[ht]
\centering
\begin{tabularx}{\textwidth}{lX}
\hline
\textbf{Field} & \textbf{Value} \\
\hline
\(\displaystyle \text{query}\) & \textit{who is the first british governor in india?} \\
\hline
\(\displaystyle \text{gold answer}\) & William Bentinck, Warren Hastings \\
\hline
\(\displaystyle \text{doc id}\) & kilt-100w\_10189364 (English) \\
\hline
\(\displaystyle P_{\text{translated}}\) & \textit{Sir John Child, 1st Baronet. Sir John Child, 1st Baronet (died 1690) was a governor of Bombay, and de facto (although not officially) the first governor-general of the British settlements in India. Born in London, Child was sent as a young boy to his uncle, the chief of the factory at Rajapur. On 27 October 1681, he was appointed chief of the East India Company's affairs at Surat and Bombay, while at the same time his namesake—stated to be unrelated by the Oxford Dictionary of National Biography—Sir Josiah Child, was governor of the company at home. The two} \\
\hline
\(\displaystyle P_{\text{refined}}\) & \textit{Sir John Child, 1st Baronet was the first governor of Bombay, and de facto (although not officially) the first governor-general of the British settlements in India. He was born in London but later moved to India where he became involved in the East India Company’s affairs. On 27th October 1681, he took up the position of Chief of the East India Company’s affairs at both Surat and Bombay. This made him the first British Governor in \ldots} \\
\hline
\end{tabularx}
\caption{A failure case of DKM-RAG because of preference of retriever so that high-resource but irrelevant document is retrieved. }
\label{tab:failure_dkmrag}
\end{table*}

\section{Language Preference of Generators in average}
\begin{figure*}[ht]
  \centering
  \includegraphics[width=1.0\textwidth]{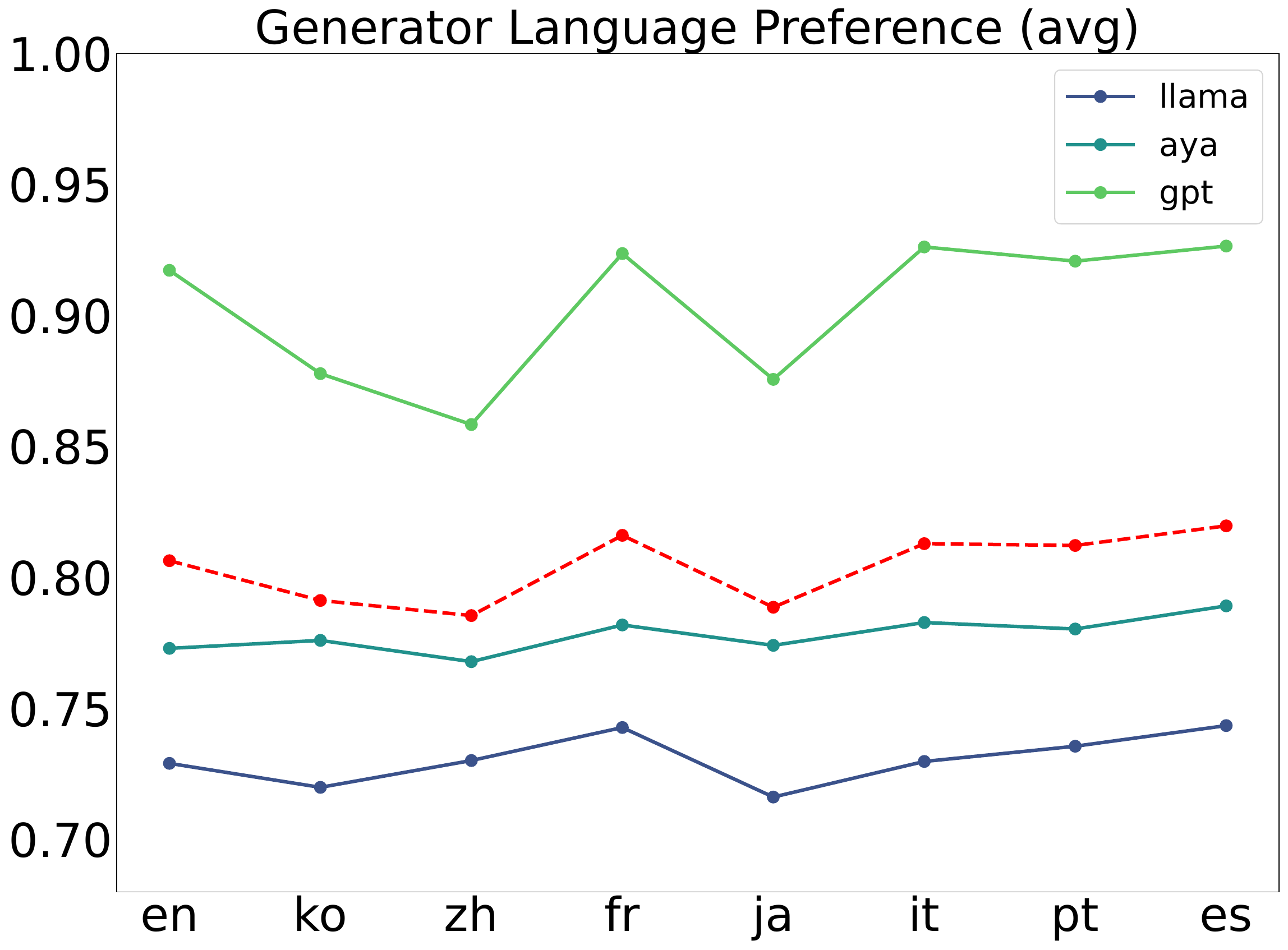}
  \caption{Average Generator Preference for three query languages: en, zh, ko.} 
  \label{fig:avg_generator_pref}
\end{figure*}

We provide language preference of generators in terms of average as shown in Figure~\ref{fig:avg_generator_pref}. Consistent with the result of each query language in Figure~\ref{fig:multi_lang}, the generator shows preferences for Latin-script languages. And GPT-4o-mini shows more consistent outputs than other generators. This is because it is a larger model than the others, providing more stable answers regardless of language preference. Between Llama and Aya, Aya produces slightly more consistent outputs, demonstrating its multilingual capability in handling diverse linguistic contexts.

\section{MLRS Analysis}
We prove the effectiveness of our proposed language preference metric, MLRS by comparing language preference between MLRS score and the average document language ratio of retrieved documents for each dataset. As stated in Table~\ref{tab:analysis_mlr}, the tendency of average language ratio of retrieved documents and MLRS score is similar. To prove it, we also report Pearson and Spearman correlation coefficients and each p-value between them. Pearson value (0.98558) indicates a very strong positive linear correlation between the average MKQA language distribution values (mkqa\_avg) and the MLRS (Preference) scores. The p-value (7.75e-10) is extremely small, showing that the probability of observing such a strong correlation by chance is almost negligible. In short, there is a statistically significant, nearly perfect linear relationship between these two sets of values. Similarly, the Spearman value (0.86264) also indicates a strong association, and the corresponding p-value (1.47e-4) confirms that this correlation is statistically significant. By these results, we prove that MLRS is efficient for measuring language preference of retriever.
\clearpage
\begin{table*}[t]
\centering
\footnotesize
\renewcommand{\arraystretch}{1.1}
\setlength{\tabcolsep}{4pt}
\begin{tabular}{lccccccccccccc}
\toprule
                & en    & ko    & ar    & zh    & fi    & fr    & de    & ja    & it    & pt    & ru    & es    & th    \\
\midrule
mkqa\_en       & 44.12 & 1.60  & 1.19  & 1.30  & 2.54  & 10.03 & 6.90  & 1.44  & 8.32  & 7.67  & 4.85  & 9.90  & 0.13  \\
mkqa\_ko       & 23.07 & 17.35 & 1.99  & 4.81  & 2.04  & 7.90  & 5.96  & 10.36 & 6.16  & 5.06  & 6.85  & 6.85  & 1.58  \\
mkqa\_ar       & 24.93 & 3.30  & 15.29 & 4.07  & 2.10  & 8.30  & 6.53  & 6.64  & 6.80  & 5.71  & 7.78  & 7.65  & 0.89  \\
mkqa\_zh       & 24.70 & 3.17  & 1.76  & 23.22 & 2.01  & 7.47  & 6.17  & 6.27  & 6.08  & 5.24  & 6.37  & 7.27  & 0.27  \\
mkqa\_fi       & 30.32 & 2.27  & 1.63  & 2.33  & 7.92  & 11.11 & 8.20  & 3.78  & 8.77  & 7.18  & 6.51  & 9.42  & 0.58  \\
mkqa\_fr       & 29.90 & 1.48  & 1.25  & 1.55  & 2.50  & 21.44 & 6.96  & 2.06  & 9.40  & 7.96  & 4.77  & 10.55 & 0.19  \\
mkqa\_de       & 32.54 & 1.46  & 1.17  & 1.44  & 2.96  & 11.40 & 15.12 & 1.89  & 9.09  & 7.69  & 4.83  & 10.17 & 0.24  \\
mkqa\_ja       & 24.56 & 4.80  & 1.69  & 3.99  & 2.19  & 7.97  & 5.99  & 22.55 & 6.38  & 5.66  & 6.49  & 7.45  & 0.28  \\
mkqa\_it       & 28.72 & 1.59  & 1.30  & 1.58  & 2.52  & 12.30 & 6.97  & 1.95  & 17.46 & 8.47  & 5.26  & 11.70 & 0.17  \\
mkqa\_pt       & 28.82 & 1.71  & 1.40  & 1.63  & 2.60  & 11.92 & 6.74  & 2.23  & 10.24 & 13.78 & 5.38  & 13.33 & 0.24  \\
mkqa\_ru       & 27.02 & 2.53  & 1.92  & 1.98  & 2.45  & 8.83  & 6.44  & 2.71  & 7.36  & 6.24  & 23.83 & 8.43  & 0.26  \\
mkqa\_es       & 29.45 & 1.73  & 1.27  & 1.60  & 2.66  & 11.85 & 6.93  & 1.83  & 10.55 & 9.33  & 5.27  & 17.36 & 0.16  \\
mkqa\_th       & 32.39 & 3.10  & 2.10  & 2.96  & 2.53  & 10.00 & 7.40  & 4.43  & 8.06  & 7.43  & 6.80  & 9.70  & 3.10  \\
\midrule
\rowcolor{gray!20} \textbf{mkqa\_avg}      & \textbf{29.27} & \textbf{3.55}  & \textbf{2.61}  & \textbf{4.04}  & \textbf{2.85}  & \textbf{10.81} & \textbf{7.41}  & \textbf{5.24}  & \textbf{8.82}  & \textbf{7.49}  & \textbf{7.31}  & \textbf{9.98}  & \textbf{0.62}  \\
\rowcolor{gray!20} \textbf{MLRS (Preference)} & \textbf{47.70} & \textbf{35.47} & \textbf{35.59} & \textbf{35.90} & \textbf{35.13} & \textbf{37.94} & \textbf{37.20} & \textbf{37.59} & \textbf{37.66} & \textbf{37.15} & \textbf{37.99} & \textbf{37.97} & \textbf{34.09} \\
\midrule
\multicolumn{14}{l}{\textbf{Pearson correlation coefficient:} 0.98558 (p-value: 7.75e-10)} \\
\multicolumn{14}{l}{\textbf{Spearman correlation coefficient:} 0.86264 (p-value: 1.47e-4)} \\
\bottomrule
\end{tabular}
\caption{Language distribution ratios of documents retrieved from datasets composed of each query language. The table lists the raw MKQA language distribution values (without the percent sign) for each dataset. The row \textbf{mkqa\_avg} shows the average distribution across all MKQA datasets for each language, while the row \textbf{MLRS (Preference)} provides the corresponding MLRS scores. Additionally, we report Pearson and Spearman correlation coefficients between MLRS and mkqa\_avg.}
\label{tab:analysis_mlr}
\end{table*}

\clearpage

\end{document}